\pdfoutput=1

\documentclass[11pt]{article}

\usepackage[]{acl}

\usepackage{times}
\usepackage{latexsym}
\usepackage{booktabs}       
\usepackage{amsfonts}       
\usepackage{colortbl}
\usepackage{bbm}
\usepackage{nicefrac} 
\usepackage{latexsym}
\usepackage{multirow}
\usepackage{booktabs}
\usepackage{amsmath}
\usepackage{booktabs}
\usepackage{subfig}
\usepackage{graphicx}
\usepackage{xspace}
\usepackage{times}
\usepackage{latexsym}
\usepackage{multirow}
\usepackage{algorithm,algorithmicx,algpseudocode}
\usepackage{url}
\usepackage{xcolor}
\usepackage[font={footnotesize}]{caption}
\usepackage{times}  
\usepackage{helvet} 
\usepackage{courier}  
\usepackage{booktabs}       
\usepackage{amsfonts}       
\usepackage{colortbl}
\usepackage{pgfplots}
\usepackage{amssymb}
\usepackage{pifont}
\usepackage{siunitx} 
\usepackage{titlesec}

\pgfplotsset{compat=1.17} 

\titlespacing{\paragraph}{%
  0pt}{
  0.0\baselineskip}{
  1em}

\usepackage[T1]{fontenc}

\usepackage[utf8]{inputenc}
\DeclareUnicodeCharacter{0307}{}
\usepackage{microtype}
\DeclareMathOperator*{\argmax}{argmax}
\definecolor{myyellow}{rgb}{0.88, 0.54, 0.35}
\definecolor{mygreen}{rgb}{0.68, 0.9, 0.8}
\definecolor{cyan}{rgb}{122, 87, 138}
\definecolor{myblue}{rgb}{0.9, 0.1, 0.94}

\iffalse
\newcommand{\pfliu}[1]{}

\newcommand{\review}[1]{}

\newcommand{\yl}[1]{}
\newcommand{\gn}[1]{}
\newcommand{\dr}[1]{}

\else
\newcommand{\pfliu}[1]{\textcolor{myblue}{\bf\small [#1 --pfliu]}}

\newcommand{\review}[1]{\textcolor{myyellow}{\bf\small [#1 --reviewer2]}}

\newcommand{\yl}[1]{\textcolor{blue}{\bf\small [#1 --yl]}}
\newcommand{\gn}[1]{\textcolor{magenta}{\bf\small [#1 --GN]}}
\newcommand{\dr}[1]{\textcolor{mygreen}{\bf\small [#1 --DR]}}
\fi

\newcommand{\model}{BRIO\xspace}
\title{BRIO: Bringing Order to Abstractive Summarization}

\author{Yixin Liu$^\textbf{1}$, Pengfei Liu$^\textbf{2}$, Dragomir Radev$^\textbf{1}$, Graham Neubig$^\textbf{2}$ \\
$^1$Yale University, $^2$Carnegie Mellon University \\
\texttt{\{yixin.liu,dragomir.radev\}@yale.edu,\{pliu3,gneubig\}@cs.cmu.edu}
  }

\begin{document}

\maketitle
\begin{abstract}

Abstractive summarization models are commonly trained using maximum likelihood estimation, which assumes a \textit{deterministic} (one-point) target distribution in which an ideal model will assign all the probability mass to the reference summary.
This assumption may lead to performance degradation during inference, where the model needs to compare several system-generated (candidate) summaries that have deviated from the reference summary.
To address this problem, we propose a novel training paradigm which assumes a \textit{non-deterministic} distribution so that different candidate summaries are assigned probability mass according to their quality. 
Our method achieves a new state-of-the-art result on the CNN/DailyMail (47.78 ROUGE-1) and XSum (49.07 ROUGE-1) datasets.
Further analysis also shows that our model can estimate probabilities of candidate summaries that are more correlated with their level of quality.\footnote{We have made our code, results, and trained models publicly available at \url{https://github.com/yixinL7/BRIO}.}

\end{abstract}

\section{Introduction}

Neural methods for abstractive summarization~\cite{rush-etal-2015-neural, nallapati-etal-2016-abstractive, chopra-etal-2016-abstractive, lewis-etal-2020-bart, zhang2020pegasus} formulate summarization as a sequence-to-sequence (Seq2Seq) problem~\citep{10.5555/2969033.2969173}, learning to generate the summary in an autoregressive manner.
Such models are commonly trained with maximum likelihood estimation (MLE), maximizing predictive probability of the reference output given the gold sub-sequence before it.
However, during inference the model must also generate the output based on possibly erroneous previous steps. This can hurt model performance, a phenomenon often called \textit{exposure bias}~\citep{10.5555/2969239.2969370, DBLP:journals/corr/RanzatoCAZ15}.
To maintain reasonable performance even in the case of a sub-sequence with errors, we argue that the model must accurately estimate relative quality of different generated outputs, since effective inference requires comparison among these candidates. 

\begin{table}[t!]
 \setlength{\tabcolsep}{9pt}
\centering \small
\begin{tabular}{lcccr}
\toprule
\textbf{System} & \textbf{R-1} & \textbf{R-2} & \textbf{R-L} & \textbf{Acc.(\%)} \\
\midrule
 High & 53.99 & 29.85 & 51.12 & 100.00 \\
 Low & 33.48 & 10.85 & 30.45 & 0.00 \\
\midrule
 BART & 44.88 & 21.68 & 41.92 & 54.80 \\
 Ours & \textbf{50.10} & \textbf{26.29} & \textbf{47.19} & \textbf{79.63} \\
\bottomrule
\end{tabular}
\caption{\label{tab:intro} 
Accuracy of different abstractive summarization systems w.r.t ranking the quality of candidate summaries on \texttt{CNNDM} dataset.
\textbf{Acc.} stands for the frequency of the model assigning higher probabilities to better candidate summaries. 
The candidate summaries are generated by a pre-trained model (BART), and we select the best and the worst candidates (w.r.t. ROUGE scores) for each of the samples.
\textbf{High} and \textbf{Low} represent the average performance of the best and worst candidates respectively.
R-1/2/L are the ROUGE-1/2/L scores.
The original BART only achieves 54.80\% accuracy.
}
\end{table}

To understand whether existing models can accurately perform such relative comparisons, we conducted a preliminary study on pre-trained BART~\citep{lewis-etal-2020-bart}, first generating two candidate summaries from the model and observing whether a higher probability is assigned to the candidate with a higher ROUGE~\citep{lin-2004-rouge} score.
As Tab.~\ref{tab:intro} shows, the accuracy is far from ideal.
This is likely due to the fact that MLE training only encourages the model to assign high probability to the reference summary, and is agnostic about any relative comparison between non-reference summaries.
However, we argue that it is also important for the order of model scores to be \textbf{coordinated} with the actual quality metrics by which the summaries will be evaluated -- higher model scores should indicate better quality summaries. In the following we will refer to models that have such scores as ``coordinated'' for conciseness.

\begin{figure}[t!]
    \centering
    \includegraphics[width=0.95\linewidth]{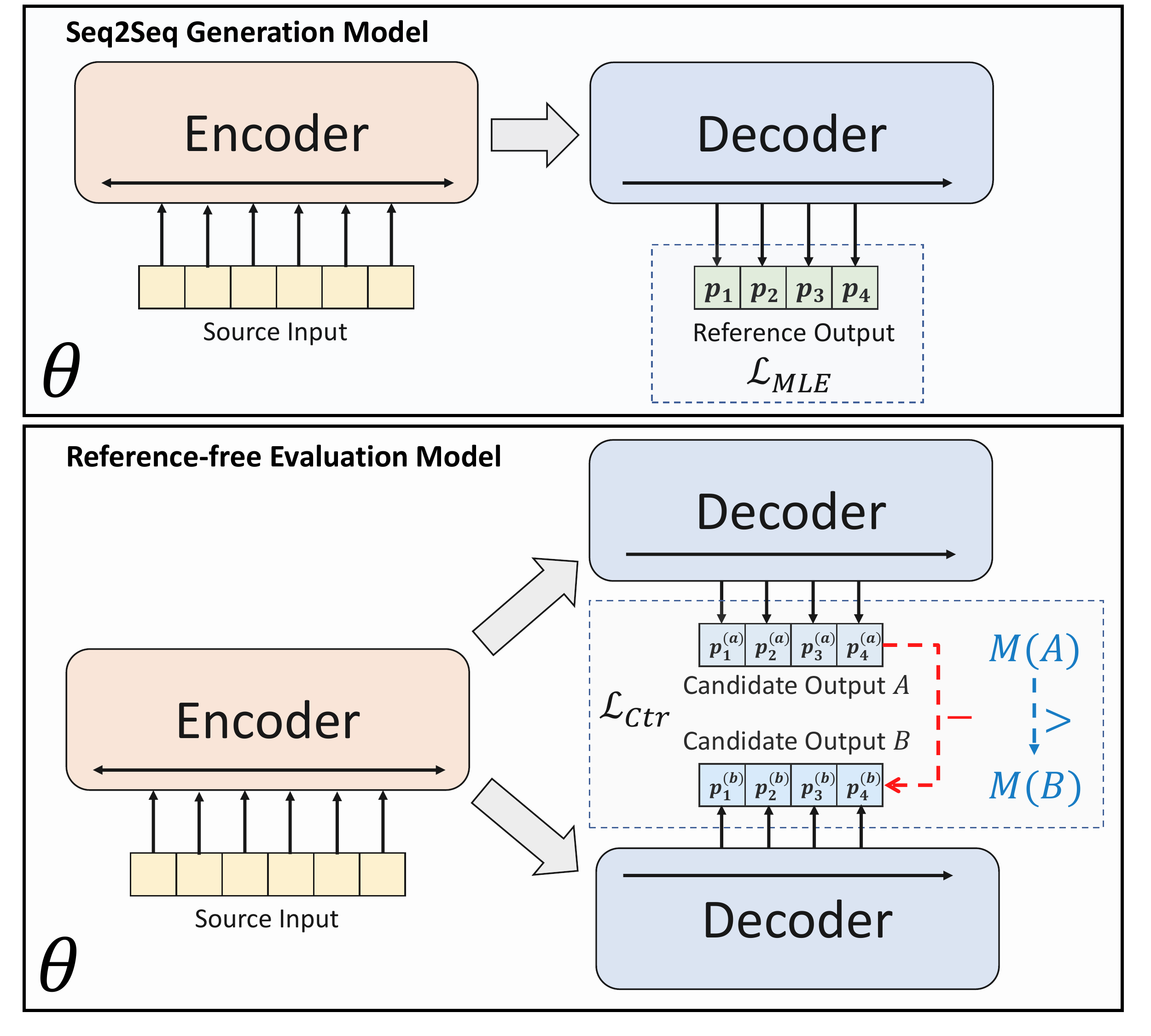}
    \caption{Comparison of MLE loss ($\mathcal{L}_{MLE}$) and the contrastive loss ($\mathcal{L}_{Ctr}$) in our method.
    MLE assumes a \textbf{deterministic} (one-point) distribution, in which the reference summary receives all the probability mass.
    Our method assumes a \textbf{non-deterministic} distribution in which system-generated summaries also receive probability mass according to their quality.
    The contrastive loss encourages the order of model-predicted probabilities of candidate summaries to be coordinated with the actual quality metric $M$ by which the summaries will be evaluated.
    We assign the abstractive model a \textit{dual} role -- a single model could be used both as a \textbf{generation} model and a reference-free \textbf{evaluation} model.}
    \label{fig:intro}
\end{figure}

We introduce a training paradigm which requires the abstractive model to be able to be \textbf{accurate} with respect to predicting the tokens in the \textit{reference} summaries and \textbf{coordinated} with respect to the \textit{candidate} summaries.
In other words, we give the abstractive model a \textit{dual} role: as a \emph{generation} model, it generates the output summaries in an autoregressive way; as an \emph{evaluation} model, it can be used to score the quality of candidate summaries by estimating a probability distribution over candidate outputs.
The generation model is trained using the standard MLE loss, but to train the evaluation model we introduce a \textit{contrastive} loss~\citep{10.1109/CVPR.2006.100} defined over different candidate summaries generated by pre-trained abstractive models (Fig.~\ref{fig:intro}), following previous work on ranking-based or contrastive learning \citep{hopkins-may-2011-tuning, zhong-etal-2020-extractive, liu-etal-2021-refsum}.

Our main contribution is to change the target distribution of abstractive models from a \textit{one-point deterministic} distribution assumed by MLE training to a \textit{non-deterministic} distribution in which candidate summaries are also assigned probability mass according to their quality.
The new SOTA performance on CNN/DailyMail~\citep{10.5555/2969239.2969428} and XSum~\citep{narayan-etal-2018-dont} datasets demonstrated the effectiveness of our method.
Our in-depth analysis also found that the abstractive models trained using our method can estimate the candidate summary quality more accurately, in concert with the the objective of our training paradigm.

\section{Neural Abstractive Summarization}

The goal of abstractive summarization is to create a function $g$ that takes a source document $D$ and generates an appropriate summary $S$
\begin{equation}
    S \leftarrow g(D)
\end{equation}

\paragraph{Training Objective}

Neural abstractive summarization models aim to learn a neural model $g$ that results in good summaries.
Maximum likelihood estimation (MLE) is the standard training algorithm. It aims to maximize the likelihood of the reference summary $S^*$, i.e.,
\begin{equation}
\label{eq:mle}
    \theta^* = \argmax_\theta \sum_i \log p_{g_\theta} ({S^*}^{(i)}|D^{(i)}; \theta)
\end{equation}
where $\theta$ denotes the parameters of $g$ and $p_{g_\theta}$ denotes the probability distribution entailed by these parameters. 
The summation is over the training set and $\{D^{(i)}, {S^*}^{(i)}\}$ is the $i$-th training sample.

For a specific sample $\{D^{(i)}, {S^*}^{(i)}\}$, Eq.~\ref{eq:mle} is equivalent to minimizing the sum of negative log-likelihoods of the tokens $\{s_1^*, \cdots , s_j^*, \cdots , s_l^*\}$ in the reference summary $S^*$ whose length is $l$, which is the cross-entropy loss:
\begin{equation}
\small
\begin{split}
    & \mathcal{L}_{xent} =  \\
    & - \sum_{j=1}^l \sum_s p_{\mathrm{true}} (s|D, S_{<j}^*) \log p_{g_\theta} (s |D, S_{<j}^*; \theta) \\
\end{split}
\label{eq:xent}
\end{equation}
where $S_{<j}^*$ denotes the partial reference sequence $\{s_0^*, \cdots, s_{j - 1}^*\}$ and $s_0^*$ is a pre-defined start token.
$p_{\mathrm{true}}$ is a one-hot distribution under the standard MLE framework:
\begin{equation}
    \small
    p_{\mathrm{true}} (s|D, S_{<j}^*) = 
    \begin{cases}
    1 & s = s_j^* \\
    0 & s \neq s_j^* \\
    \end{cases}
\label{eq:one-hot}
\end{equation}

In practice, label smoothing~\citep{7780677} is a widely used and effective technique that modifies the target distribution in Eq.~\ref{eq:one-hot} to a "soft" label by assigning probability mass $\beta$ to other tokens:
\begin{equation}
    \small
    p_{\mathrm{true}} (s|D, S_{<j}^*) = 
    \begin{cases}
    1 - \beta & s = s_j^* \\
    \frac{\beta}{N-1} & s \neq s_j^* \\
    \end{cases}
\label{eq:label_smoothing}
\end{equation}
where $N$ is the size of the dictionary.

\paragraph{Inference and Exposure Bias}

During inference, the abstractive model $g$ is used to generate the candidate summary in an autoregressive manner.
It is intractable to enumerate all the possible candidate outputs, so in practice methods such as \textit{beam search} are used to reduce the search space.

One important step in search is estimating the probability of the next word $s_t$ given the previous predicted sequence $S_{<t}$:
\begin{equation}
\label{eq:beam}
    p_{g_\theta}(s_{t}|D, S_{<t}; \theta)
\end{equation}
Comparing Eq.~\ref{eq:beam} with Eq.~\ref{eq:xent}, the major difference is that during inference the model makes new predictions based on its \textbf{own} previous predictions $S_{<t}$ instead of the \textit{reference} $S^*_{<t}$.
As a result, even if the generation model $g$ achieves very high accuracy w.r.t.~Eq.~\ref{eq:xent}, once $S_{<t}$ starts to deviate from $S^*$, there is the risk that the performance of $g$ will significantly degrade.
This problem has been identified as the \textit{exposure bias}~\citep{10.5555/2969239.2969370}.

\section{Coordinating Abstractive Models}

Eq.~\ref{eq:beam} implies that the abstractive model $g$ should be able to assign higher estimated probability to the better candidate summary during inference.
However, this intuition is not directly captured in the standard MLE objective used in training -- a model obtaining zero MLE loss would assign zero probability to any candidate summary different from the reference.
This is obviously improper for any task where multiple reasonable generations may exist \citep{khayrallah-etal-2020-simulated}, and also does not say anything about the ordering of two imperfect references.
We therefore advocate for making the alternative assumption that the probability of one candidate should be well-correlated with its quality as evaluated by an automatic metric $M$.
Since it is intractable to enumerate all the possible candidate outputs, we only require our model to be able to accurately predict the ranking order of a set of the most probable candidate summaries $\hat{\mathcal{S}}$, which are its own beam search results.
In order to achieve this objective, we slightly modify the conditions of Eq.~\ref{eq:label_smoothing}, maintaining the general functional form, but instead specifying the \emph{marginal} probability of the non-reference candidates $\mathcal{S}$ to be $\beta$, and encouraging \emph{coordination} of probabilities and qualities among non-reference candidates as follows:
\begin{equation}
\small
 \begin{cases}
    p_{\mathrm{true}^\dag} (S|D) = 1 - \beta & S = S^* \\
    \sum_{S \in \mathcal{S}} p_{\mathrm{true}^\dag} (S|D) = \beta & S \neq S^* \\
   \multirow{2}{*}{$p_{\mathrm{true}^\dag} (S_i|D) > p_{\mathrm{true}^\dag} (S_j|D)$}  & \forall S_i, S_j \in \hat{\mathcal{S}}, \\
      & M(S_i) > M(S_j)  \\
\end{cases}
\label{eq:dist_ours}
\end{equation}
We next describe precisely how we encourage coordination through \emph{contrastive learning}.
\paragraph{Contrastive Learning for Coordination}
The candidate quality measure $M$ can be defined in many ways. 
In this work we define it as the ROUGE~\citep{lin-2004-rouge} score of a candidate summary $S_i$ given the reference summary $S^*$.
To coordinate a pre-trained abstractive model, we 1) use it to generate different candidate summaries with various levels of quality,\footnote{This is achieved by using diverse beam search~\citep{Vijayakumar_Cogswell_Selvaraju_Sun_Lee_Crandall_Batra_2018}.}
then 2) encourage the model to assign higher estimated probabilities to better candidates by fine-tuning the model with a \textbf{contrastive} loss, following the previous work~\citep{hopkins-may-2011-tuning, zhong-etal-2020-extractive}:
\begin{equation}
\small
\label{eq:contr}
      \mathcal{L}_{ctr} = \sum_i \sum_{j > i} \max(0, f(S_j) - f(S_i) + \lambda_{ij})
\end{equation}
where $S_i$ and $S_j$ are two different candidate summaries and $\textrm{ROUGE}(S_i, S^*) > \textrm{ROUGE}(S_j, S^*)$, $ \forall i, j, i < j$.
$\lambda_{ij}$ is the margin multiplied by the difference in rank between the candidates, i.e., $\lambda_{ij} = (j - i) * \lambda $. 
$f(S_i)$ is the length-normalized estimated log-probability\footnote{We length-normalize as it is standard in comparing hypotheses in neural sequence generation \citep{cho-etal-2014-properties}.}
\begin{equation}
    f(S) = \frac{\sum_{t=1}^{l} \log p_{g_\theta} (s_t |D, S_{<t}; \theta)}{|S|^\alpha}
\label{eq:score}
\end{equation}
where $\alpha$ is the length penalty hyperparameter.

This loss gives the abstractive model a dual purpose, first as a reference-free \textit{evaluation} model, which can be used in a \textbf{two-stage} summarization pipeline, where it is used to score the candidates generated by a pre-trained \textit{generation} model and select the final output from them. 
However, since the autoregressive generation depends on both the \textbf{token-level prediction accuracy} and \textbf{sequence-level coordination}, the model fine-tuned with the contrastive loss alone can no longer be used as a \textit{generation} model. 

\paragraph{Multi-task Fine-tuning}

Following \citet{edunov-etal-2018-classical}, we combine the contrastive (Eq.~\ref{eq:contr}) and cross-entropy (Eq.~\ref{eq:xent}) losses to preserve the \textbf{generation} ability of the pre-trained abstractive model: 
\begin{equation}
\label{eq:mul}
    \mathcal{L}_{mul} = \mathcal{L}_{xent} + \gamma \mathcal{L}_{ctr}
\end{equation}
where $\gamma$ is the weight of the contrastive loss.
We note that the contrastive and the cross-entropy loss can effectively complement each other -- since the contrastive loss is defined on the sequence level, the token-level cross-entropy loss serves as a normalization to ensure that the model could assign balanced probability mass across the whole sequence. 

\section{Related Work}

\paragraph{Training Methods of Seq2Seq Models}
\label{subsec:training_method}
In order to align the training objective and evaluation metric, structured losses have been used for the Seq2Seq model training. 
Among them, margin-based losses~\citep{ Herbrich99supportvector, NIPS2003_878d5691, gimpel-smith-2010-softmax}, which require the model to assign higher probability to the better output, are a major category.
Many margin-based losses used in modern seq2seq models~\citep{ wiseman-rush-2016-sequence, edunov-etal-2018-classical} assume a deterministic (one-point) distribution: a model can achieve zero loss if it can assign a much higher probability to the (pseudo)-reference, regardless of relative comparisons of other candidate summaries.
By contrast, our method has a non-deterministic assumption (Eq.~\ref{eq:dist_ours}), which focuses on the pair-wise ranking of a set of candidate summaries.

One main challenge of directly optimizing a Seq2Seq model with quality scores of the output is that the discrete sampling process makes the loss non-differentiable. 
To circumvent this problem, reinforcement learning has been used to reformulate the conditional text generation tasks~\citep{ DBLP:journals/corr/RanzatoCAZ15, DBLP:journals/corr/BahdanauBXGLPCB16, li-etal-2016-deep, paulus2018a, li-etal-2019-deep}.
Compared to this school of methods, our method is based on supervised learning, and it is more stable and less sensitive to the design choices (e.g. reward shaping), which are well-known challenges of reinforcement learning methods.
Minimum risk training~\citep{shen-etal-2016-minimum, wieting-etal-2019-beyond} and other online sampling based methods~\citep{10.5555/2969239.2969370, NIPS2016_2f885d0f, zhang-etal-2019-bridging} belong to another school of methods used to circumvent the problem of non-differentiability. However, they also exhibit similar problems of stability as reinforcement learning.

\paragraph{Contrastive Learning}

Recently, contrastive learning~\citep{10.1109/CVPR.2006.100} has been introduced into several conditional text generation tasks, such as machine translation~\citep{yang-etal-2019-reducing, pan-etal-2021-contrastive}, text summarization~\citep{cao-wang-2021-cliff, DBLP:journals/corr/abs-2109-03481, DBLP:journals/corr/abs-2108-11846}, and other tasks~\citep{uehara-etal-2020-learning,cho-etal-2021-contrastive, lee2021contrastive}. 
Among these application scenarios, most work deployed contrastive learning in the \textit{latent representation space}, following the framework proposed in \citet{pmlr-v119-chen20j}.
However, in this work we adopt contrastive learning over the \textit{discrete} space of the generated texts.

Besides, instead of constructing the contrastive learning examples by rule-based methods (e.g. perturbing the reference output), we use the generation models to construct the examples, which makes the contrastive learning task closer to the generation task.
\citet{DBLP:journals/corr/abs-2108-11846} also adopted contrastive learning on the generated texts. 
However, their formulation belongs to the margin-based losses.
We have discussed the difference between our method and the margin-based losses in the previous paragraphs.

\noindent \textbf{Discriminative Reranking}
Discriminative reranking has been widely studied for conditional generation tasks~\citep{shen-etal-2004-discriminative, och-etal-2004-smorgasbord,Wan2015MultiDocumentSV, mizumoto-matsumoto-2016-discriminative}.
Some recent works~\citep{liu-liu-2021-simcls, lee-etal-2021-discriminative} have also explored discriminative reranking of candidates from neural natural language generation models, which adopt large pre-trained language models (e.g. BERT~\citep{devlin-etal-2019-bert}) as the reranker.
In this work, we factorize the Seq2Seq model (e.g., BART) trained on the same dataset as the reranking model, which maximizes the parameter sharing across two stages.
Besides, our approach contributes an instance of leveraging large pre-trained Seq2Seq models as a quality estimation model~\citep{yuan2021bartscore}.

\section{Experiments}

\subsection{Experimental Settings}

\paragraph{Datasets} We mainly use three datasets in our experiments (statistics in Appendix~\ref{sec:datasets}).

\noindent\texttt{CNNDM}\footnote{\url{https://cs.nyu.edu/~kcho/DMQA/}}~\citep{10.5555/2969239.2969428} is a large scale news dataset.
Following \citet{nallapati-etal-2016-abstractive}, we treat the news articles as the source documents and the associated highlights as the summaries.

\noindent\texttt{XSum}\footnote{\url{https://github.com/EdinburghNLP/XSum}}~\citep{narayan-etal-2018-dont} is a highly abstractive dataset of articles from the British Broadcasting Corporation (BBC).

\noindent\texttt{NYT\footnote{\url{https://catalog.ldc.upenn.edu/LDC2008T19}}}~\citep{linguistic2008new}
contains articles from the New York Times and the associated summaries.
We follow \citet{kedzie-etal-2018-content} for data preprocessing and splitting, and use the associated archival abstracts as the summaries.

\paragraph{Baselines}

We choose a variety of related models with strong performance as baselines.
\noindent \textbf{BART}~\citep{lewis-etal-2020-bart} and \textbf{PEGASUS}~\citep{zhang2020pegasus} are both large pre-trained Seq2Seq LMs standard in the literature.
\textbf{GSum}~\citep{dou-etal-2021-gsum} is built on BART, and improves performance by using additional guidance from an extractive summarizer.
\textbf{SimCLS}~\citep{liu-liu-2021-simcls} introduces a two-stage framework where the pre-trained BART model is used to generate candidates and a pre-trained RoBERTa~\citep{DBLP:journals/corr/abs-1907-11692} model is fine-tuned as an evaluation model to score the candidate summaries and select from them. It achieves state-of-the-art performance on both \texttt{CNNDM} and \texttt{XSum}.
\textbf{GOLD}~\citep{pang2021text} uses offline reinforcement learning to train the BART model by treating the reference summaries as the demonstrations, a different formulation that can also improve the performance of the original BART.
\textbf{SeqCo}~\citep{DBLP:journals/corr/abs-2109-03481} and \textbf{ConSum}~\citep{DBLP:journals/corr/abs-2108-11846} are two recent methods that aim to leverage contrastive learning to improve the performance of the abstractive summarization model (BART). 

\paragraph{Implementation Details}

In the following experiments, we use either BART or PEGASUS as a backbone.
We label our proposed methods {\textbf{\model}}, with two variants: (1) {\textbf{\model-Ctr}} is fine-tuned with the contrastive loss (Eq.~\ref{eq:contr}) only; (2) {\textbf{\model-Mul}} is fine-tuned with the multi-task loss (Eq.~\ref{eq:mul}).
We use {\textbf{\model-Ctr}} as an evaluation model that scores different candidate summaries generated by a Seq2Seq abstractive model and selects the final output from them, and \textbf{\model-Mul} as a standard Seq2Seq model that takes the source documents as input and generates the output in an autoregressive manner. 
Further details are in Appendix~\ref{sec:appendix}.

\subsection{Results}

The results are shown in Tab~\ref{tab:other}.
For \texttt{CNNDM} and \texttt{NYT} we use BART as the backbone model while for \texttt{XSum} we use the pre-trained PEGASUS model as our base model since it achieves better performance than BART.
We have the following observations:

(1) \model-Ctr outperforms \textit{SimCLS}, its counterpart as an evaluation model in a two-stage summarization framework.
Specifically, both \model-Ctr and SimCLS are used to score the candidate summaries generated by a Seq2Seq abstractive model (BART).
The final outputs are selected based on those scores.
We attribute \model-Ctr's superior performance to its use of the same model architecture (BART) for both candidate generation and scoring, while SimCLS uses RoBERTa as the evaluation model.
As a result, \model-Ctr maximizes the parameter sharing between the two stages, and preserves the power of the Seq2Seq model pre-trained on the same dataset.

\begin{table}[t!]
\centering
\small
\begin{tabular}{lccc}
\toprule
\textbf{System} & \textbf{R-1} & \textbf{R-2} & \textbf{R-L}  \\
\midrule
\multicolumn{4}{c}{CNNDM} \\
\midrule
 BART* & 44.16 & 21.28 & 40.90 \\
 
 PEGASUS* & 44.17 & 21.47 & 41.11  \\
 GSum* & 45.94 & 22.32 & 42.48 \\
 ConSum* & 44.53 & 21.54 & 41.57 \\
 SeqCo* & 45.02 & 21.80 & 41.75 \\
 GOLD-\textit{p}* & 45.40 & 22.01 & 42.25 \\
 GOLD-\textit{s}* & 44.82 & 22.09 & 41.81 \\
 SimCLS* & 46.67 & 22.15 & 43.54 \\
 $\textrm{BART}^\ddag$ & 44.29 & 21.17 & 41.09 \\
\midrule
 \model-Ctr & $47.28^\dag$ & $22.93^\dag$ & $44.15^\dag$ \\
 \model-Mul & $\textbf{47.78}^\dag$ & $\textbf{23.55}^\dag$ & $\textbf{44.57}^\dag$ \\
\midrule
\multicolumn{4}{c}{XSum} \\
\midrule
 BART* & 45.14 & 22.27 & 37.25 \\
 PEGASUS* & 47.21 & 24.56 & 39.25 \\
 GSum* & 45.40 & 21.89 & 36.67 \\
 ConSum* & 47.34 & 24.67 & 39.40 \\
 SeqCo* & 45.65 & 22.41 & 37.04 \\
 GOLD-\textit{p}* & 45.75 & 22.26 & 37.30 \\
 GOLD-\textit{s}* & 45.85 & 22.58 & 37.65 \\
 SimCLS* & 47.61 & 24.57 & 39.44 \\
 $\textrm{PEGASUS}^\ddag$ & 47.46 & 24.69 & 39.53 \\
\midrule
 \model-Ctr & $48.13^\dag$ & $25.13^\dag$ & $39.84^\dag$ \\
 \model-Mul & $\textbf{49.07}^\dag$ & $\textbf{25.59}^\dag$ & $\textbf{40.40}^\dag$ \\
\midrule
\multicolumn{4}{c}{NYT} \\
\midrule
 $\textrm{BART}^\ddag$  & 55.78 & 36.61 & 52.60 \\
\midrule
\model-Ctr & 55.98 & 36.54 & 52.51 \\
 \model-Mul & $\textbf{57.75}^\dag$ & $\textbf{38.64}^\dag$ & $\textbf{54.54}^\dag$ \\
\bottomrule
\end{tabular}
\caption{\label{tab:other} Results on \texttt{CNNDM}, \texttt{XSum} and \texttt{NYT}.
On \texttt{NYT} we only reported our own results due to different data pre-processing.
\dag: significantly better than the baseline model ($p < 0.01$).
*: results reported in the original papers.
\ddag: results from our own evaluation script.
R-1/2/L are the ROUGE-1/2/L F$_1$ scores.}
\end{table}

(2) \model-Mul is able to establish the new stare-of-the-art performance on \texttt{CNNDM}.
Notably, the previous state-of-the-art model, \textit{GSum}, takes additional guidance as input and needs a separate encoder to encode the guidance information, while \model-Mul uses the same parameterization of BART.
Compared to other methods (ConSum, SeqCo, GOLD) that aim to improve upon BART, \model-Mul performs much better, showing the effectiveness of our training method.

(3) Since on \texttt{XSum} we use PEGASUS instead of BART as the base model, the result shows that our method is not restricted to the specific choice of the base model.

\subsection{Analysis}
\label{subsec:analysis}

\begin{table}[t!]
\centering
\small
\begin{tabular}{lccc}
\toprule
\textbf{Coefficient ($\gamma$)} & \textbf{R-1} & \textbf{R-2} & \textbf{R-L}  \\
\midrule
 0 (BART) & 44.29 & 21.17 & 41.09 \\
 0.1 & 45.08  & 21.63  & 41.71  \\
 1 & 46.01 & 22.22 & 42.68  \\
 2 & 46.36 & 22.79 & 43.07 \\
 5 & 46.91 & 23.03 & 43.63 \\
 10 & 47.22 & 23.31 & 43.94 \\
 100 & \textbf{47.78} & \textbf{23.55} & \textbf{44.57} \\
 1000 & 46.83 & 22.17 & 43.68 \\
 $+\infty$ \small(\model-Ctr) & 47.28 & 22.93 & 44.15 \\
\bottomrule
\end{tabular}
\caption{\label{tab:coeff} Model performance with different $\gamma$ coefficients weighting the contrastive loss (Eq.~\ref{eq:mul}) on \texttt{CNNDM}. 
\model-Ctr is trained with the contrastive loss only, which no longer preserves its generation ability. 
We report its performance when it is used as an evaluation model to select from candidate summaries.
R-1/2/L are the ROUGE-1/2/L F$_1$ scores. }
\end{table}

We further perform some in-depth analyses from diverse perspectives on the \texttt{CNNDM} dataset to gain more insights into our proposed method.

\paragraph{Coefficients of the Multi-Task Loss} The multi-task loss (Eq.~\ref{eq:mul}) used to train our model contains two parts: the cross-entropy loss and the contastive loss. 
As shown in Tab.~\ref{tab:coeff}, as the weight of the contrastive loss ($\gamma$) increases, the model's performance improves. 
However, the cross-entropy loss is still necessary to preserve the model's ability as a generation model.
We argue that this is because the token level accuracy is still important during the auto-regressive generation process, where the individual tokens are predicted sequentially. 
In addition, we also found that the model tends to achieve the best performance (w.r.t the ROUGE scores on the development set) faster with a higher $\gamma$.
Specifically, it requires less than one entire epoch to achieve the best performance on \texttt{CNNDM}, making our approach an efficient fine-tuning method.

\paragraph{Generation-Finetuning as a Loop}

\begin{figure}[t!]
    \centering
    \includegraphics[width=0.8\linewidth]{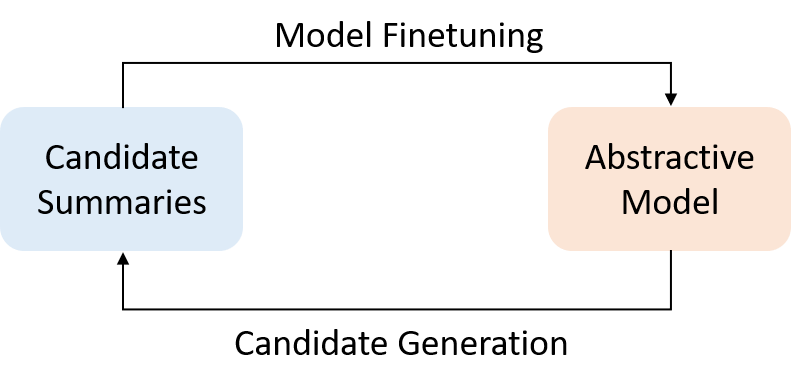}
    \caption{Loop of candidate generation and model finetuning. }
    \label{fig:loop}
\end{figure}

\begin{table}[t!]
\centering
\small
\begin{tabular}{lccc}
\toprule
\textbf{System} & \textbf{R-1} & \textbf{R-2} & \textbf{R-L}  \\
\midrule
 BART & 44.29 & 21.17 & 41.09 \\
 \model-Mul & 47.78 & 23.55 & 44.57 \\
 \midrule
 \model-Loop & $\textbf{48.01}^\dag$ & $\textbf{23.80}^\dag$ & $\textbf{44.67}^\dag$ \\
\bottomrule
\end{tabular}
\caption{\label{tab:loop} Results on \texttt{CNNDM} when the pre-trained model are fine-tuned twice.
\textbf{\model-Loop} is trained on the candidates generated by \textbf{\model-Mul}.
\dag: significantly better than the baseline (BART) ($p < 0.01$).
R-1/2/L are ROUGE-1/2/L F$_1$ scores.}
\end{table}

Since the fine-tuned model (\model-Mul) is still able to generate, we can use it to generate a new set of candidates in the same way as we used the pre-trained BART model, and continue fine-tuning it on this newly created set of candidates \citep{och-2003-minimum}. 
Fig.~\ref{fig:loop} illustrates this iterative process.
The results shown in Tab.~\ref{tab:loop} illustrate that this new model (\model-Loop) outperforms \model-Mul.
Besides, the model reached the best performance very quickly, showing the potential of adopting our method in an online framework where the new candidates are dynamically generated from the current model.
We leave this direction for future work.

\paragraph{Increasing the Beam Width}

While theoretically a larger beam width (i.e. the number of candidates maintained during beam search) would allow more candidates to be considered and therefore increase the upper bound of the performance, in practice model performance may be lower if the beam width is too large.
The reason for this phenomenon is closely related to the low sequence-level coordination of the generator.
Specifically, increasing the beam width may introduce candidates with lower quality \citep{stahlberg-byrne-2019-nmt}, and the generator may not be able to differentiate them from high-quality candidates.

\begin{table}[t!]
\centering
\small
\begin{tabular}{lccccc}
\toprule
 \textbf{Beams} & \multicolumn{2}{c}{\textbf{BART}} & & \multicolumn{2}{c}{\textbf{\model-Mul}} \\
 \midrule
 & \textbf{R-1} & \textbf{R-2} & & \textbf{R-1} & \textbf{R-2} \\
\cmidrule{2-3} \cmidrule{5-6}

 4 & \textbf{44.29} & \textbf{21.17} &  & 47.78 & 23.55  \\
 10 & 43.83 & 20.76 &  & 47.98 & 23.81  \\
 20 & 43.53 & 20.49 & & 48.07 & 23.92  \\
 50 & 43.06 & 20.05 &  & 48.18 & 24.01 \\
 100 & 42.79 & 19.76 &  & \textbf{48.23} & \textbf{24.09} \\
\bottomrule
\end{tabular}
\caption{\label{tab:num_beams} Results on \texttt{CNNDM} with different beam widths (the number of beams) used in beam search.
The default beam width is 4.
R-1/2 are the ROUGE-1/2 F$_1$ scores.
}
\end{table}

In Tab.~\ref{tab:num_beams}, we compare the performance of the pre-trained BART and our model (\model-Mul) with different beam widths used during inference.
We observe that the performance of BART goes down as the beam width increases.
On the other hand, our model is able to achieve better performance with a larger number of beams, demonstrating that our training method can improve the coordination of the model by encouraging the model to assign estimated probabilities to candidate summaries well-correlated with their quality.

\paragraph{Training with Different Evaluation Metrics}

\begin{table}[t!]
\centering
\small
\begin{tabular}{lcccc}
\toprule
\textbf{System} & \textbf{R-1} & \textbf{R-2} & \textbf{R-L} & \textbf{BS} \\
\midrule
 BART & 44.29 & 21.17 & 41.09 & 27.38\\
 \model-Mul (R) & \textbf{47.78} & \textbf{23.55} & \textbf{44.57} & 32.11\\
 \model-Mul (B) & 47.53 & 23.22 & 44.37 & \textbf{32.59} \\
\bottomrule
\end{tabular}
\caption{\label{tab:bertscore} Results on \texttt{CNNDM} using different evaluation metrics as $M$ in Eq.\ref{eq:dist_ours}.
\model-Mul (R) is trained with candidate summaries ordered by ROUGE scores, while \model-Mul (B) is trained with candidate summaries ordered by BERTScore.
R-1/2/L are ROUGE-1/2/L F$_1$ scores.
BS denotes BERTScore.}
\end{table}

In the previous experiments, we used ROUGE as the evaluation metric to define the target ordering of the candidate summaries (Eq.\ref{eq:dist_ours}).
To evaluate our method's performance beyond ROUGE, we use a model-based semantic similarity metric, BERTScore~\citep{Zhang*2020BERTScore:},\footnote{\url{https://github.com/Tiiiger/bert_score}. We use its default version for English texts.} as the evaluation metric $M$ in Eq.\ref{eq:dist_ours} to compare the performance of different candidate summaries.
Then, we trained another version of \model-Mul based on the order of candidate summaries calculated by BERTScore.

The results in Tab.~\ref{tab:bertscore} show that (1)
Our model can significantly improve the model performance when either ROUGE or BERTScore is used as the target evaluation metric for ordering candidate summaries. 
This suggests that it is possible to use our method to optimize any specific target metric, making our method an alternative to reinforcement learning or minimum risk training. 
(2) Our model that is trained on one evaluation metric (e.g. BERTScore) also achieves improvement on another metric (e.g. ROUGE) compared with the baseline model, which indicates that the improvement made by our model is not from exploiting the potential weaknesses of individual metrics.
Besides, this result also demonstrates a non-trivial degree of agreement between ROUGE and BERTScore.

\begin{table}[t!]
\centering
\small
\begin{tabular}{lcc}
\toprule
\textbf{System} & \textbf{Unigram} & \textbf{Bigram} \\
\midrule
 Reference & .1110 & .4865  \\
\midrule
 BART & .0101 & .0924 \\
 \model-Mul & .0262 & .2381 \\
\bottomrule
\end{tabular}
\caption{\label{tab:novel} Ratio of novel $n$-grams of different models on \texttt{CNNDM}.
Novel $n$-grams are those that appear in the summaries but not in the source documents.}
\end{table}

\begin{figure}[t!]
    \centering
    \includegraphics[width=1\linewidth]{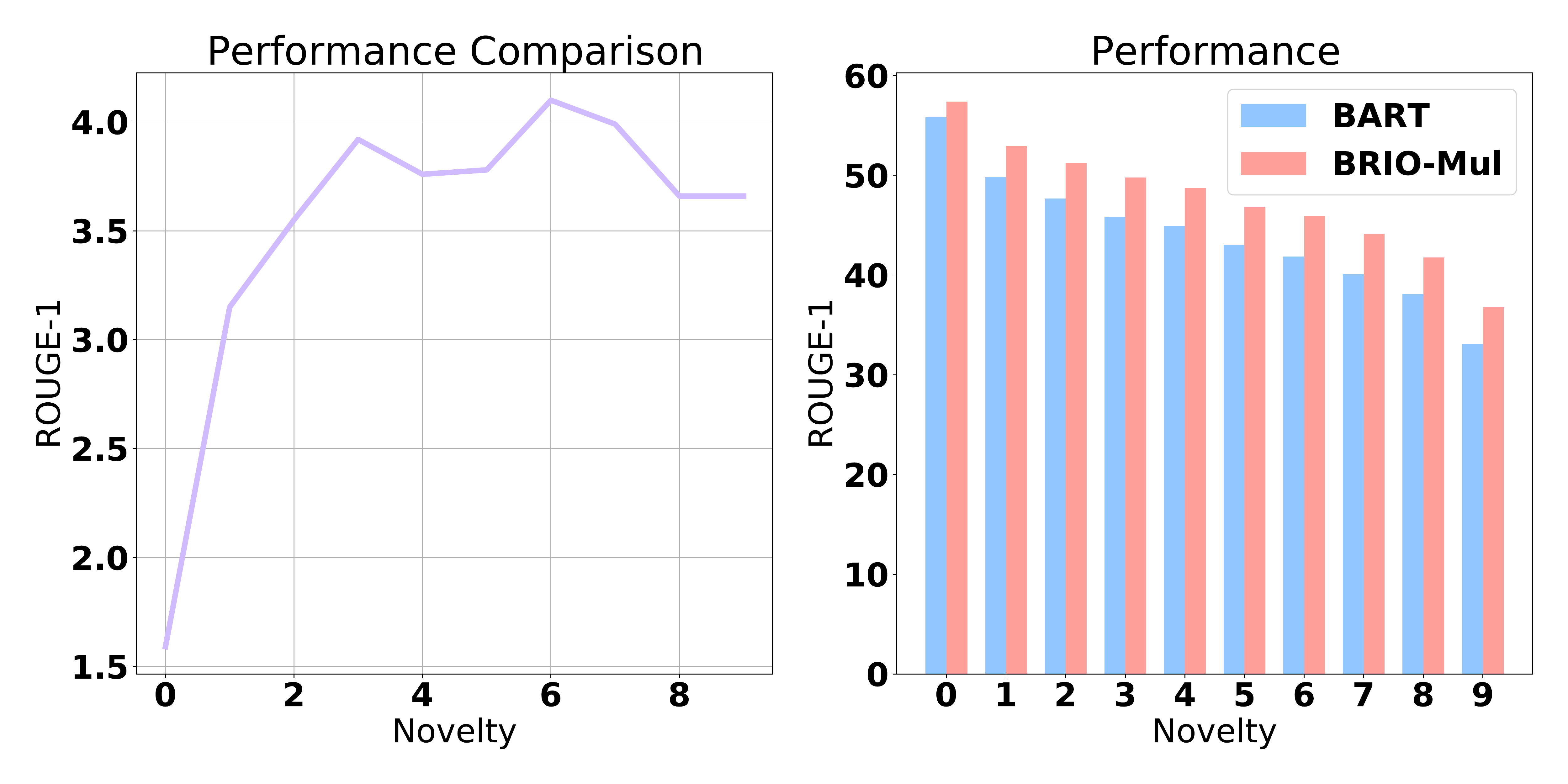}
    \caption{Performance comparison (BART v.s. \model-Mul) w.r.t. reference summary novelty. The x-axis represents different buckets of test examples grouped by reference summary novelty (Eq.~\ref{eq:novel}). Larger x-coordinates correspond to examples of which the reference summaries have higher novelty. 
    The left figure shows the performance improvement of our model compared with the baseline model, while the right one shows model performance.}
    \label{fig:novelty}
\end{figure}

\paragraph{Novel $n$-grams} We compare the ratio of novel $n$-grams in reference, \model-Mul's, and BART's summaries.
As Tab.~\ref{tab:novel} shows, our model is more ``abstractive'' compared to BART, although reference summaries still contain more novel $n$-grams.
This is likely due to the fact that our model is optimized at the sequence-level, allowing more freedom for paraphrasing and compression.

We further investigate the relation of the ``abstractiveness" and model performance by comparing our model (\model-Mul) with the baseline model (BART) on different buckets of test examples grouped by the ``novelty" of the reference summaries,\footnote{The calculation is performed using ExplainaBoard~\citep{liu-etal-2021-explainaboard}. \url{https://github.com/neulab/ExplainaBoard}.} i.e.,
\begin{equation}
\label{eq:novel}
\small
    \mathrm{Novelty}(D, S^*) = \frac{\sum_{g \in G_{S^*}}\mathbbm{1}(g \notin G_D)}{|G_{S^*}|} 
\end{equation}
where $D$ and $S^*$ are the source document and reference summary respectively, $G_D$ and $G_{S^*}$ are the sets of bigrams in $D$ and $S^*$, $\mathbbm{1}$ is the indicator function.
The results in Fig.~\ref{fig:novelty} show that when novelty is higher, (1) all models' performance decreases; (2) our model achieves larger improvement over the baseline model.

\paragraph{Rank Correlation}

\begin{table}[t!]
\centering
\small
\begin{tabular}{c|c|c}
\toprule
   & \textbf{Own} & \textbf{PEGASUS} \\
 \midrule
\textbf{BART} & .0470  & .1205  \\
\midrule
\textbf{\model-Mul} & $\mathbf{.1839}^\dag$  & $\mathbf{.2768}^\dag$ \\
\bottomrule
\end{tabular}
\caption{\label{tab:rank_correlation} Rank Correlation between the model's estimated probabilities of the candidate summaries and the quality scores (ROUGE) of the candidate summaries on \texttt{CNNDM}.
\textbf{Own} stands for the candidates generated by the models themselves, while \textbf{PEGASUS} stands for the candidates generated by the pre-trained PEGASUS model.
\dag: significantly better than the baseline model (BART) ($p < 0.01$).
}
\end{table}

We computed the rank correlation between the \textbf{estimated probabilities} of the candidate summaries calculated by the generators and the \textbf{quality scores} of the candidate summaries. 
We use Eq.~\ref{eq:score} to calculate the estimated probabilities\footnote{We found the value of the length penalty factor $\alpha$ in Eq.~\ref{eq:score} by maximizing the rank correlation on the validation set.} and we use ROUGE-1 as the quality score metric of the candidate summaries.
We calculate Spearman's rank correlation for each sample, and use the average score as the overall correlation, 

We investigated two specific settings: 1) ranking candidate summaries generated by a different model (PEGASUS); 2) ranking candidate summaries generated by themselves (BART \& \model-Mul).
We use 16 candidates in total for calculation. 
As Tab.~\ref{tab:rank_correlation} shows, our model achieves better rank correlation on the candidate summaries generated by both itself and the independent model.
This suggests that our model can better estimate the quality of candidate summaries.

\subsection{Token-level Calibration}

\begin{table}[t]
\centering
\small
\begin{tabular}{llccc}
\toprule
\textbf{Dataset} & \textbf{System} & \textbf{ECE} & \textbf{Acc} & \textbf{Conf}\\
\midrule
 \multirow{2}{*}{CNNDM} & BART & .4097 & .3711 & .7365\\
  & \model-Mul & \textbf{.2719} & .4271 & .6652\\
\midrule
 \multirow{2}{*}{XSum} & PEGASUS & .2369 & .4688 & .6990\\
  & \model-Mul &  \textbf{.1423} & .4744  & .5881\\
\bottomrule
\end{tabular}
\caption{\label{tab:ece} Expected Calibration Error (ECE), accuracy (Acc) and confidence (Conf) on the test set of \texttt{CNNDM} and \texttt{XSum}.}
\end{table}

\begin{figure}[t]
    \centering
    \includegraphics[width=0.98\linewidth]{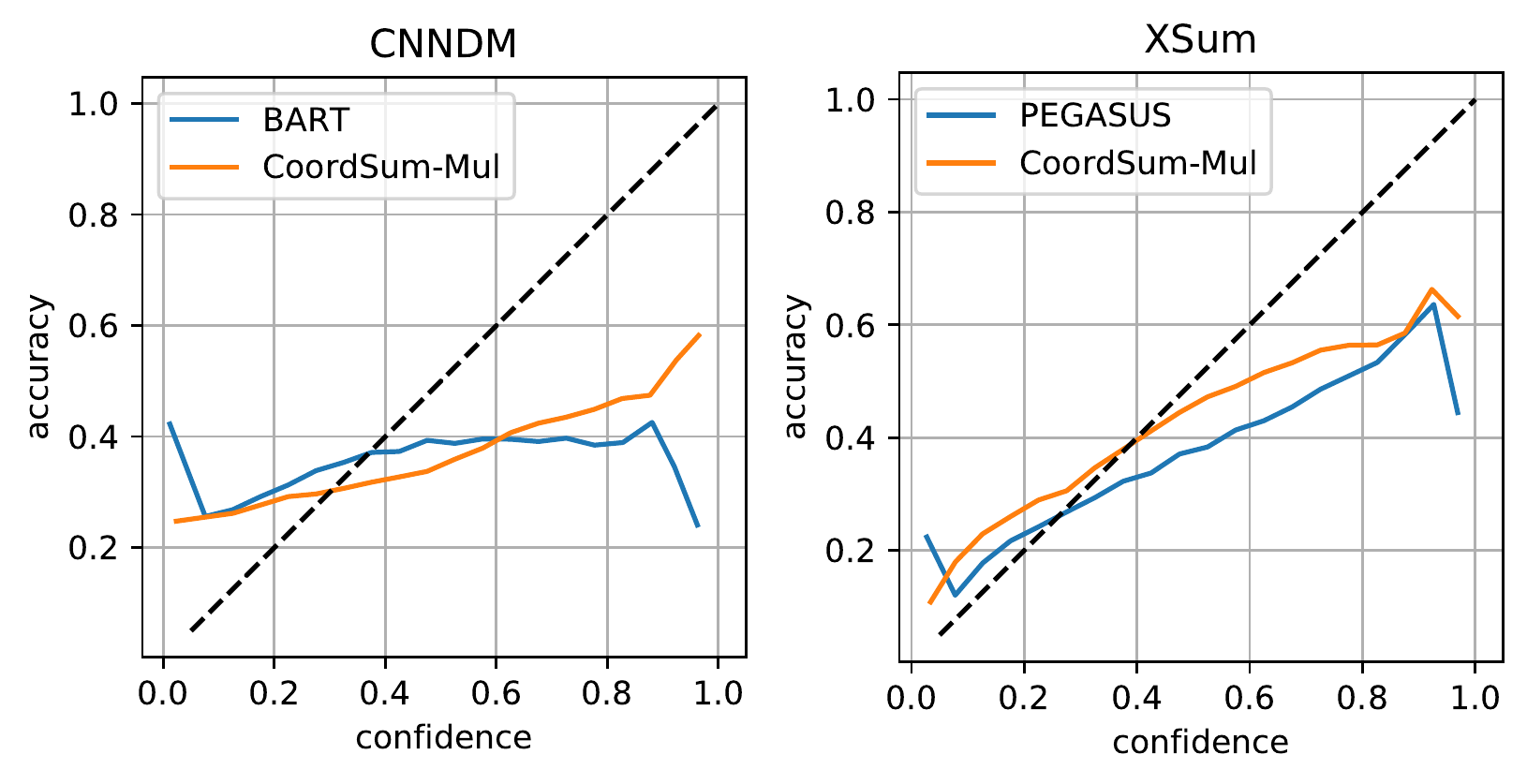}
    \caption{Reliability graphs on the \texttt{CNNDM} and \texttt{XSum} datasets.
    The accuracy of model's predictions is plotted against the model's confidence on these predictions.}
    \label{fig:ece}
\end{figure}

Calibration requires that a model's confidence on its predictions is equal to the accuracy of these predictions~\citep{pmlr-v70-guo17a}.
Previous work~\citep{NEURIPS2019_f1748d6b, DBLP:journals/corr/abs-1903-00802, wang-etal-2020-inference} has found that a more \textit{calibrated} text generation model tends to have better performance, and 
techniques like \textit{label smoothing} can improve both the \textbf{token-level} calibration and \textbf{sequence-level} accuracy (i.e. the ability of generating better results).
One intuitive explanation of this phenomenon is to interpret the model's estimated probability of a generated summary as the product of the model's confidences on a series of token-level predictions.
Then, since a more calibrated model's \textit{confidence} estimates better the \textit{accuracy} of its predictions, the model's estimated \textbf{probability} of one sequence should be more indicative of the \textbf{quality} of this sequence, which is essential for the beam search during inference. 
However, the relation of token-level calibration and sequence-level performance remains inconclusive~\citep{NEURIPS2019_f1748d6b}.\footnote{In general, better token-level calibration doesn't guarantee better sequence-level performance.} 
For example, a generator that always predicts a uniform distribution over all tokens would be perfectly calibrated, however, such a model would not generate high-quality outputs.

We investigate this relation from the opposite direction by evaluating whether our model (\model-Mul), which is trained to have better \textbf{sequence-level} performance, would also be more calibrated at the \textbf{token-level} compared with the baseline models that are trained using MLE and label smoothing.
We follow previous work by using the \textit{Expected Calibration Error}~\citep{10.5555/2888116.2888120} (ECE) as the evaluation metric of calibration:
\begin{equation}
    \textrm{ECE} = \sum_{m=1}^M \frac{|B_m|}{n} | \textrm{acc}(B_m) - \textrm{conf}(B_m) |
\end{equation}
where the samples are grouped into $M$ equal-width buckets by confidence (conf), $B_m$ denotes the $m$-th bucket, and $n$ is the total number of samples.
Following \citet{wang-etal-2020-inference}, we evaluate model calibration on the \textbf{system-generated} summaries during inference and use the tercom toolkit\footnote{\url{http://cs.umd.edu/~snover/tercom/}} 
to assign labels (correct/incorrect) to the system-generated summaries based on the reference summaries.

\begin{table*}[t!]
    \scriptsize
    \centering
    \extrarowheight=\aboverulesep
    \addtolength{\extrarowheight}{\belowrulesep}
    \aboverulesep=0pt
    \belowrulesep=0pt
    \begin{tabular}{@{} c  p{0.85\textwidth}}
     \toprule
    \multicolumn{1}{l}{ \bf System} &  \multicolumn{1}{c}{ \bf Summary} \\\midrule
     \multicolumn{1}{l}{ \bf Reference} & \cellcolor{gray!25} chelsea forward tammy abraham nets first-half double for chelsea. dominic solanke adds a third late on as chelsea look set to win trophy. manchester city struggle without injured star thierry ambrose. read: mourinho warns his young chelsea players he can not play them all. \textbf{click here} to read our match report from man city 's academy stadium. \\
     \multicolumn{1}{l}{ \bf BART} & tammy abraham scored twice in the first half to give chelsea the lead. isaac buckley-ricketts levelled the game for manchester city. dominic solanke scored late on to put a gloss on the scoreline. \textbf{click here} to read sportsmail's player ratings from the youth cup final. \\
    \multicolumn{1}{l}{ \bf \model-Mul }& \cellcolor{gray!25} chelsea beat manchester city 3-1 in the youth cup final at the etihad stadium. tammy abraham scored twice in the first half to give chelsea the lead. dominic solanke scored late on to seal the win for the home side. \\
   
  \midrule
   \multicolumn{1}{l}{ \bf Reference} & \cellcolor{gray!25} alejandro valverde won ahead of julian alaphilippe and michael albasini. chris froome finished 123rd after a crash during the final 12 kilometres. team sky's sports director gabriel rasch praised froome for finishing. rasch said froome was `banged up' but expects to ride tour de romandie. \\
    \multicolumn{1}{l}{ \bf BART }&  movistar rider alejandro valverde won fleche wallonne on wednesday. team sky's chris froome fell in the final 12km but finished the race. philippe gilbert pulled out of the race after a bad crash 50km from the end. \textbf{click here} for more cycling news. \\
    \multicolumn{1}{l}{ \bf \model-Mul} & \cellcolor{gray!25} alejandro valverde defended his fleche wallonne title in belgium on wednesday. movistar rider finished ahead of julian alaphilippe and michael albasini. team sky's chris froome fell in the final 12km of the race but finished in 123rd. froome was involved in a crash but finished the race despite being `banged up' \\
\midrule
\multicolumn{1}{l}{ \bf Reference} & \cellcolor{gray!25} manuel pellegrini won the premier league and capital one cup last season. city currently sit fourth in the league table - 12 points behind chelsea. pellegrini's contract expires at the end of the 2015-16 season. city players have been impressed with vieira's work with the youth team. pep guardiola is city's first-choice to succeed pellegrini at the etihad. \\
   \multicolumn{1}{l}{ \bf BART} &  manuel pellegrini's future at manchester city is under scrutiny. patrick vieira is highly-respected among the city players. city's first-choice managerial option is bayern munich boss pep guardiola. \textbf{click here} for all the latest manchester city news. \textbf{click here} for more premier league news. \\
    \multicolumn{1}{l}{ \bf \model-Mul }& \cellcolor{gray!25} manchester city players have backed patrick vieira to replace manuel pellegrini as manager of the club. the frenchman is highly-respected among the players at the etihad stadium. pellegrini's future at the club is under scrutiny after a disappointing season. city's first-choice manager is current bayern munich boss pep guardiola. \\
    
  \bottomrule
\end{tabular}
\vspace{-5pt}
\caption{Case Study on \texttt{CNNDM}. \model-Mul learns to ignore the noise pattern (``click here") while BART cannot.}
\vspace{-8pt}
\label{tab:example}
\end{table*}

The results in Tab.~\ref{tab:ece} show that \model-Mul is better calibrated compared to BART, suggesting that our method helps to improve the token-level calibration by explicitly encouraging the model to have more accurate sequence-level probability estimations.
The reliability graph is shown in Fig.~\ref{fig:ece}.
We found that (1) abstractive models are generally over-confident on their own predictions, 
(2) models are generally more calibrated on \texttt{XSum} than \texttt{CNNDM}.
This is likely due to the fact that \texttt{XSum} has shorter summaries therefore it is less likely to be affected by the exposure bias.

\subsection{Few-shot Fine-tuning}

\begin{table}
\centering
\small
\begin{tabular}{lcccc}
\toprule
\textbf{Dataset} & \textbf{System} & \textbf{R-1} & \textbf{R-2} & \textbf{R-L}  \\
\midrule
 \multirow{2}{*}{CNNDM} & BART & 44.29 & 21.17 & 41.09 \\
& \model-Few & \textbf{45.81} & \textbf{21.91} & \textbf{42.61} \\
\midrule
 \multirow{2}{*}{XSum} & PEGASUS & 47.46 & 24.69 & 39.53\\
& \model-Few & \textbf{47.95} & \textbf{24.89} & \textbf{39.71} \\
\bottomrule
\end{tabular}
\vspace{-5pt}
\caption{\label{tab:few-shot} Few-shot Fine-tuning.
\textbf{\model-Few} is trained on only 100/1000 training examples on \texttt{CNNDM} and \texttt{XSum} respectively.
R-1/2/L are ROUGE-1/2/L F$_1$ scores.}
\vspace{-8pt}
\end{table}

The training paradigm proposed in this paper may be extended to any Seq2Seq model. 
However, it can be a non-trivial overhead to generate the candidate summaries using large neural models on the entire training set.
On the other hand, recent work~\citep{JMLR:v21:20-074, zhang2020pegasus, schick-schutze-2021-shot, fabbri-etal-2021-improving} has shown that few-shot learning can be an effective fine-tuning method of pre-trained models for text generation tasks.

Therefore, we investigate our model's performance in a few-shot setting.
Specifically, we randomly sample 100/1000 examples from the training set of \texttt{CNNDM}/\texttt{XSum}, and fine-tune the models that are pre-trained using MLE loss on those examples.
More training details can be found in Appendix~\ref{appendix:few-shot}.
The results are shown in Tab.~\ref{tab:few-shot}.
All experiments are repeated three times, and the reported results are the average performance. 
The results indicate that our model can achieve improvement over the baseline model under the few-shot learning setting with a small computational overhead.

\subsection{Case Study on CNNDM}

Tab.~\ref{tab:example} presents an interesting pattern we observed when comparing the results of \model-Mul and BART, which demonstrates that our method helps the abstractive model to filter out noise patterns in the original data.
Specifically, some of the reference summaries (331/11490) in \texttt{CNNDM} contains the phrase ``click here'', pointing to a hyperlink, and 103 source documents also contain this phrase.
BART picked up this pattern, and generates this phrase in 96 output summaries.
On the contrary, our model learns to ignore this noise pattern and never generated it across the whole test set, likely because it identified that generated candidates with this pattern rarely achieve a high ROUGE score, and downweighted the probability accordingly.

\section{Conclusion and Future Work}

In this work, we presented a new training paradigm that assigns candidate outputs probability mass according to their quality using contrastive learning.
While our method has achieved significant improvement on abstractive summarization, we note several directions for the future work to explore.
First, since our method makes no assumptions specifically about the summarization task, it can be extended to other conditional text generation tasks such as machine translation.
Second, it is possible to apply  our method in a reinforcement learning setting, where the candidate summaries are dynamically generated.
Finally, in experiments we only used diverse beam search to generate the candidate summaries, but it is likely that other candidate generation methods could yield further improvements.

\section*{Acknowledgements}
We thank the anonymous reviewers for valuable feedback and helpful suggestions.

\bibliography{anthology,custom}

\begin{thebibliography}{68}
\expandafter\ifx\csname natexlab\endcsname\relax\def\natexlab#1{#1}\fi

\bibitem[{Bahdanau et~al.(2016)Bahdanau, Brakel, Xu, Goyal, Lowe, Pineau,
  Courville, and Bengio}]{DBLP:journals/corr/BahdanauBXGLPCB16}
Dzmitry Bahdanau, Philemon Brakel, Kelvin Xu, Anirudh Goyal, Ryan Lowe, Joelle
  Pineau, Aaron~C. Courville, and Yoshua Bengio. 2016.
\newblock \href {http://arxiv.org/abs/1607.07086} {An actor-critic algorithm
  for sequence prediction}.
\newblock \emph{CoRR}, abs/1607.07086.

\bibitem[{Bengio et~al.(2015)Bengio, Vinyals, Jaitly, and
  Shazeer}]{10.5555/2969239.2969370}
Samy Bengio, Oriol Vinyals, Navdeep Jaitly, and Noam Shazeer. 2015.
\newblock Scheduled sampling for sequence prediction with recurrent neural
  networks.
\newblock In \emph{Proceedings of the 28th International Conference on Neural
  Information Processing Systems - Volume 1}, NIPS'15, page 1171–1179,
  Cambridge, MA, USA. MIT Press.

\bibitem[{Cao and Wang(2021)}]{cao-wang-2021-cliff}
Shuyang Cao and Lu~Wang. 2021.
\newblock \href {https://aclanthology.org/2021.emnlp-main.532} {{CLIFF}:
  Contrastive learning for improving faithfulness and factuality in abstractive
  summarization}.
\newblock In \emph{Proceedings of the 2021 Conference on Empirical Methods in
  Natural Language Processing}, pages 6633--6649, Online and Punta Cana,
  Dominican Republic. Association for Computational Linguistics.

\bibitem[{Chen et~al.(2020)Chen, Kornblith, Norouzi, and
  Hinton}]{pmlr-v119-chen20j}
Ting Chen, Simon Kornblith, Mohammad Norouzi, and Geoffrey Hinton. 2020.
\newblock \href {https://proceedings.mlr.press/v119/chen20j.html} {A simple
  framework for contrastive learning of visual representations}.
\newblock In \emph{Proceedings of the 37th International Conference on Machine
  Learning}, volume 119 of \emph{Proceedings of Machine Learning Research},
  pages 1597--1607. PMLR.

\bibitem[{Cho et~al.(2014)Cho, van Merri{\"e}nboer, Bahdanau, and
  Bengio}]{cho-etal-2014-properties}
Kyunghyun Cho, Bart van Merri{\"e}nboer, Dzmitry Bahdanau, and Yoshua Bengio.
  2014.
\newblock \href {https://doi.org/10.3115/v1/W14-4012} {On the properties of
  neural machine translation: Encoder{--}decoder approaches}.
\newblock In \emph{Proceedings of {SSST}-8, Eighth Workshop on Syntax,
  Semantics and Structure in Statistical Translation}, pages 103--111, Doha,
  Qatar. Association for Computational Linguistics.

\bibitem[{Cho et~al.(2021)Cho, Zhang, Rao, Celikyilmaz, Xiong, Gao, Wang, and
  Dolan}]{cho-etal-2021-contrastive}
Woon~Sang Cho, Yizhe Zhang, Sudha Rao, Asli Celikyilmaz, Chenyan Xiong,
  Jianfeng Gao, Mengdi Wang, and Bill Dolan. 2021.
\newblock \href {https://aclanthology.org/2021.eacl-main.2} {Contrastive
  multi-document question generation}.
\newblock In \emph{Proceedings of the 16th Conference of the European Chapter
  of the Association for Computational Linguistics: Main Volume}, pages 12--30,
  Online. Association for Computational Linguistics.

\bibitem[{Chopra et~al.(2016)Chopra, Auli, and
  Rush}]{chopra-etal-2016-abstractive}
Sumit Chopra, Michael Auli, and Alexander~M. Rush. 2016.
\newblock \href {https://doi.org/10.18653/v1/N16-1012} {Abstractive sentence
  summarization with attentive recurrent neural networks}.
\newblock In \emph{Proceedings of the 2016 Conference of the North {A}merican
  Chapter of the Association for Computational Linguistics: Human Language
  Technologies}, pages 93--98, San Diego, California. Association for
  Computational Linguistics.

\bibitem[{Devlin et~al.(2019)Devlin, Chang, Lee, and
  Toutanova}]{devlin-etal-2019-bert}
Jacob Devlin, Ming-Wei Chang, Kenton Lee, and Kristina Toutanova. 2019.
\newblock \href {https://doi.org/10.18653/v1/N19-1423} {{BERT}: Pre-training of
  deep bidirectional transformers for language understanding}.
\newblock In \emph{Proceedings of the 2019 Conference of the North {A}merican
  Chapter of the Association for Computational Linguistics: Human Language
  Technologies, Volume 1 (Long and Short Papers)}, pages 4171--4186,
  Minneapolis, Minnesota. Association for Computational Linguistics.

\bibitem[{Dou et~al.(2021)Dou, Liu, Hayashi, Jiang, and
  Neubig}]{dou-etal-2021-gsum}
Zi-Yi Dou, Pengfei Liu, Hiroaki Hayashi, Zhengbao Jiang, and Graham Neubig.
  2021.
\newblock \href {https://doi.org/10.18653/v1/2021.naacl-main.384} {{GS}um: A
  general framework for guided neural abstractive summarization}.
\newblock In \emph{Proceedings of the 2021 Conference of the North American
  Chapter of the Association for Computational Linguistics: Human Language
  Technologies}, pages 4830--4842, Online. Association for Computational
  Linguistics.

\bibitem[{Edunov et~al.(2018)Edunov, Ott, Auli, Grangier, and
  Ranzato}]{edunov-etal-2018-classical}
Sergey Edunov, Myle Ott, Michael Auli, David Grangier, and Marc{'}Aurelio
  Ranzato. 2018.
\newblock \href {https://doi.org/10.18653/v1/N18-1033} {Classical structured
  prediction losses for sequence to sequence learning}.
\newblock In \emph{Proceedings of the 2018 Conference of the North {A}merican
  Chapter of the Association for Computational Linguistics: Human Language
  Technologies, Volume 1 (Long Papers)}, pages 355--364, New Orleans,
  Louisiana. Association for Computational Linguistics.

\bibitem[{Fabbri et~al.(2021)Fabbri, Han, Li, Li, Ghazvininejad, Joty, Radev,
  and Mehdad}]{fabbri-etal-2021-improving}
Alexander Fabbri, Simeng Han, Haoyuan Li, Haoran Li, Marjan Ghazvininejad,
  Shafiq Joty, Dragomir Radev, and Yashar Mehdad. 2021.
\newblock \href {https://doi.org/10.18653/v1/2021.naacl-main.57} {Improving
  zero and few-shot abstractive summarization with intermediate fine-tuning and
  data augmentation}.
\newblock In \emph{Proceedings of the 2021 Conference of the North American
  Chapter of the Association for Computational Linguistics: Human Language
  Technologies}, pages 704--717, Online. Association for Computational
  Linguistics.

\bibitem[{Gimpel and Smith(2010)}]{gimpel-smith-2010-softmax}
Kevin Gimpel and Noah~A. Smith. 2010.
\newblock \href {https://aclanthology.org/N10-1112} {Softmax-margin {CRF}s:
  Training log-linear models with cost functions}.
\newblock In \emph{Human Language Technologies: The 2010 Annual Conference of
  the North {A}merican Chapter of the Association for Computational
  Linguistics}, pages 733--736, Los Angeles, California. Association for
  Computational Linguistics.

\bibitem[{Guo et~al.(2017)Guo, Pleiss, Sun, and Weinberger}]{pmlr-v70-guo17a}
Chuan Guo, Geoff Pleiss, Yu~Sun, and Kilian~Q. Weinberger. 2017.
\newblock \href {https://proceedings.mlr.press/v70/guo17a.html} {On calibration
  of modern neural networks}.
\newblock In \emph{Proceedings of the 34th International Conference on Machine
  Learning}, volume~70 of \emph{Proceedings of Machine Learning Research},
  pages 1321--1330. PMLR.

\bibitem[{Hadsell et~al.(2006)Hadsell, Chopra, and
  LeCun}]{10.1109/CVPR.2006.100}
Raia Hadsell, Sumit Chopra, and Yann LeCun. 2006.
\newblock \href {https://doi.org/10.1109/CVPR.2006.100} {Dimensionality
  reduction by learning an invariant mapping}.
\newblock In \emph{Proceedings of the 2006 IEEE Computer Society Conference on
  Computer Vision and Pattern Recognition - Volume 2}, CVPR '06, page
  1735–1742, USA. IEEE Computer Society.

\bibitem[{Herbrich et~al.(1999)Herbrich, Graepel, and
  Obermayer}]{Herbrich99supportvector}
Ralf Herbrich, Thore Graepel, and Klaus Obermayer. 1999.
\newblock Support vector learning for ordinal regression.
\newblock In \emph{In International Conference on Artificial Neural Networks},
  pages 97--102.

\bibitem[{Hermann et~al.(2015)Hermann, Ko\v{c}isk\'{y}, Grefenstette, Espeholt,
  Kay, Suleyman, and Blunsom}]{10.5555/2969239.2969428}
Karl~Moritz Hermann, Tom\'{a}\v{s} Ko\v{c}isk\'{y}, Edward Grefenstette, Lasse
  Espeholt, Will Kay, Mustafa Suleyman, and Phil Blunsom. 2015.
\newblock Teaching machines to read and comprehend.
\newblock In \emph{Proceedings of the 28th International Conference on Neural
  Information Processing Systems - Volume 1}, NIPS'15, page 1693–1701,
  Cambridge, MA, USA. MIT Press.

\bibitem[{Hopkins and May(2011)}]{hopkins-may-2011-tuning}
Mark Hopkins and Jonathan May. 2011.
\newblock \href {https://aclanthology.org/D11-1125} {Tuning as ranking}.
\newblock In \emph{Proceedings of the 2011 Conference on Empirical Methods in
  Natural Language Processing}, pages 1352--1362, Edinburgh, Scotland, UK.
  Association for Computational Linguistics.

\bibitem[{Kedzie et~al.(2018)Kedzie, McKeown, and
  Daum{\'e}~III}]{kedzie-etal-2018-content}
Chris Kedzie, Kathleen McKeown, and Hal Daum{\'e}~III. 2018.
\newblock \href {https://doi.org/10.18653/v1/D18-1208} {Content selection in
  deep learning models of summarization}.
\newblock In \emph{Proceedings of the 2018 Conference on Empirical Methods in
  Natural Language Processing}, pages 1818--1828, Brussels, Belgium.
  Association for Computational Linguistics.

\bibitem[{Khayrallah et~al.(2020)Khayrallah, Thompson, Post, and
  Koehn}]{khayrallah-etal-2020-simulated}
Huda Khayrallah, Brian Thompson, Matt Post, and Philipp Koehn. 2020.
\newblock \href {https://doi.org/10.18653/v1/2020.emnlp-main.7} {Simulated
  multiple reference training improves low-resource machine translation}.
\newblock In \emph{Proceedings of the 2020 Conference on Empirical Methods in
  Natural Language Processing (EMNLP)}, pages 82--89, Online. Association for
  Computational Linguistics.

\bibitem[{Kingma and Ba(2015)}]{DBLP:journals/corr/KingmaB14}
Diederik~P. Kingma and Jimmy Ba. 2015.
\newblock \href {http://arxiv.org/abs/1412.6980} {Adam: {A} method for
  stochastic optimization}.
\newblock In \emph{3rd International Conference on Learning Representations,
  {ICLR} 2015, San Diego, CA, USA, May 7-9, 2015, Conference Track
  Proceedings}.

\bibitem[{Kumar and Sarawagi(2019)}]{DBLP:journals/corr/abs-1903-00802}
Aviral Kumar and Sunita Sarawagi. 2019.
\newblock \href {http://arxiv.org/abs/1903.00802} {Calibration of encoder
  decoder models for neural machine translation}.
\newblock \emph{CoRR}, abs/1903.00802.

\bibitem[{Lee et~al.(2021{\natexlab{a}})Lee, Auli, and
  Ranzato}]{lee-etal-2021-discriminative}
Ann Lee, Michael Auli, and Marc{'}Aurelio Ranzato. 2021{\natexlab{a}}.
\newblock \href {https://doi.org/10.18653/v1/2021.acl-long.563} {Discriminative
  reranking for neural machine translation}.
\newblock In \emph{Proceedings of the 59th Annual Meeting of the Association
  for Computational Linguistics and the 11th International Joint Conference on
  Natural Language Processing (Volume 1: Long Papers)}, pages 7250--7264,
  Online. Association for Computational Linguistics.

\bibitem[{Lee et~al.(2021{\natexlab{b}})Lee, Lee, and
  Hwang}]{lee2021contrastive}
Seanie Lee, Dong~Bok Lee, and Sung~Ju Hwang. 2021{\natexlab{b}}.
\newblock \href {https://openreview.net/forum?id=Wga_hrCa3P3} {Contrastive
  learning with adversarial perturbations for conditional text generation}.
\newblock In \emph{International Conference on Learning Representations}.

\bibitem[{Lewis et~al.(2020)Lewis, Liu, Goyal, Ghazvininejad, Mohamed, Levy,
  Stoyanov, and Zettlemoyer}]{lewis-etal-2020-bart}
Mike Lewis, Yinhan Liu, Naman Goyal, Marjan Ghazvininejad, Abdelrahman Mohamed,
  Omer Levy, Veselin Stoyanov, and Luke Zettlemoyer. 2020.
\newblock \href {https://doi.org/10.18653/v1/2020.acl-main.703} {{BART}:
  Denoising sequence-to-sequence pre-training for natural language generation,
  translation, and comprehension}.
\newblock In \emph{Proceedings of the 58th Annual Meeting of the Association
  for Computational Linguistics}, pages 7871--7880, Online. Association for
  Computational Linguistics.

\bibitem[{Li et~al.(2016)Li, Monroe, Ritter, Jurafsky, Galley, and
  Gao}]{li-etal-2016-deep}
Jiwei Li, Will Monroe, Alan Ritter, Dan Jurafsky, Michel Galley, and Jianfeng
  Gao. 2016.
\newblock \href {https://doi.org/10.18653/v1/D16-1127} {Deep reinforcement
  learning for dialogue generation}.
\newblock In \emph{Proceedings of the 2016 Conference on Empirical Methods in
  Natural Language Processing}, pages 1192--1202, Austin, Texas. Association
  for Computational Linguistics.

\bibitem[{Li et~al.(2019)Li, Lei, Qin, and Wang}]{li-etal-2019-deep}
Siyao Li, Deren Lei, Pengda Qin, and William~Yang Wang. 2019.
\newblock \href {https://doi.org/10.18653/v1/D19-1623} {Deep reinforcement
  learning with distributional semantic rewards for abstractive summarization}.
\newblock In \emph{Proceedings of the 2019 Conference on Empirical Methods in
  Natural Language Processing and the 9th International Joint Conference on
  Natural Language Processing (EMNLP-IJCNLP)}, pages 6038--6044, Hong Kong,
  China. Association for Computational Linguistics.

\bibitem[{Lin(2004)}]{lin-2004-rouge}
Chin-Yew Lin. 2004.
\newblock \href {https://aclanthology.org/W04-1013} {{ROUGE}: A package for
  automatic evaluation of summaries}.
\newblock In \emph{Text Summarization Branches Out}, pages 74--81, Barcelona,
  Spain. Association for Computational Linguistics.

\bibitem[{Liu et~al.(2021{\natexlab{a}})Liu, Fu, Xiao, Yuan, Chang, Dai, Liu,
  Ye, and Neubig}]{liu-etal-2021-explainaboard}
Pengfei Liu, Jinlan Fu, Yang Xiao, Weizhe Yuan, Shuaichen Chang, Junqi Dai,
  Yixin Liu, Zihuiwen Ye, and Graham Neubig. 2021{\natexlab{a}}.
\newblock \href {https://doi.org/10.18653/v1/2021.acl-demo.34}
  {{E}xplaina{B}oard: An explainable leaderboard for {NLP}}.
\newblock In \emph{Proceedings of the 59th Annual Meeting of the Association
  for Computational Linguistics and the 11th International Joint Conference on
  Natural Language Processing: System Demonstrations}, pages 280--289, Online.
  Association for Computational Linguistics.

\bibitem[{Liu et~al.(2019)Liu, Ott, Goyal, Du, Joshi, Chen, Levy, Lewis,
  Zettlemoyer, and Stoyanov}]{DBLP:journals/corr/abs-1907-11692}
Yinhan Liu, Myle Ott, Naman Goyal, Jingfei Du, Mandar Joshi, Danqi Chen, Omer
  Levy, Mike Lewis, Luke Zettlemoyer, and Veselin Stoyanov. 2019.
\newblock \href {http://arxiv.org/abs/1907.11692} {Roberta: {A} robustly
  optimized {BERT} pretraining approach}.
\newblock \emph{CoRR}, abs/1907.11692.

\bibitem[{Liu et~al.(2021{\natexlab{b}})Liu, Dou, and
  Liu}]{liu-etal-2021-refsum}
Yixin Liu, Zi-Yi Dou, and Pengfei Liu. 2021{\natexlab{b}}.
\newblock \href {https://doi.org/10.18653/v1/2021.naacl-main.113} {{R}ef{S}um:
  Refactoring neural summarization}.
\newblock In \emph{Proceedings of the 2021 Conference of the North American
  Chapter of the Association for Computational Linguistics: Human Language
  Technologies}, pages 1437--1448, Online. Association for Computational
  Linguistics.

\bibitem[{Liu and Liu(2021)}]{liu-liu-2021-simcls}
Yixin Liu and Pengfei Liu. 2021.
\newblock \href {https://doi.org/10.18653/v1/2021.acl-short.135} {{S}im{CLS}: A
  simple framework for contrastive learning of abstractive summarization}.
\newblock In \emph{Proceedings of the 59th Annual Meeting of the Association
  for Computational Linguistics and the 11th International Joint Conference on
  Natural Language Processing (Volume 2: Short Papers)}, pages 1065--1072,
  Online. Association for Computational Linguistics.

\bibitem[{Mizumoto and
  Matsumoto(2016)}]{mizumoto-matsumoto-2016-discriminative}
Tomoya Mizumoto and Yuji Matsumoto. 2016.
\newblock \href {https://doi.org/10.18653/v1/N16-1133} {Discriminative
  reranking for grammatical error correction with statistical machine
  translation}.
\newblock In \emph{Proceedings of the 2016 Conference of the North {A}merican
  Chapter of the Association for Computational Linguistics: Human Language
  Technologies}, pages 1133--1138, San Diego, California. Association for
  Computational Linguistics.

\bibitem[{M\"{u}ller et~al.(2019)M\"{u}ller, Kornblith, and
  Hinton}]{NEURIPS2019_f1748d6b}
Rafael M\"{u}ller, Simon Kornblith, and Geoffrey~E Hinton. 2019.
\newblock \href
  {https://proceedings.neurips.cc/paper/2019/file/f1748d6b0fd9d439f71450117eba2725-Paper.pdf}
  {When does label smoothing help?}
\newblock In \emph{Advances in Neural Information Processing Systems},
  volume~32. Curran Associates, Inc.

\bibitem[{Naeini et~al.(2015)Naeini, Cooper, and
  Hauskrecht}]{10.5555/2888116.2888120}
Mahdi~Pakdaman Naeini, Gregory~F. Cooper, and Milos Hauskrecht. 2015.
\newblock Obtaining well calibrated probabilities using bayesian binning.
\newblock In \emph{Proceedings of the Twenty-Ninth AAAI Conference on
  Artificial Intelligence}, AAAI'15, page 2901–2907. AAAI Press.

\bibitem[{Nallapati et~al.(2016)Nallapati, Zhou, dos Santos, Gu̇l{\c{c}}ehre,
  and Xiang}]{nallapati-etal-2016-abstractive}
Ramesh Nallapati, Bowen Zhou, Cicero dos Santos, {\c{C}}a{\u{g}}lar
  Gu̇l{\c{c}}ehre, and Bing Xiang. 2016.
\newblock \href {https://doi.org/10.18653/v1/K16-1028} {Abstractive text
  summarization using sequence-to-sequence {RNN}s and beyond}.
\newblock In \emph{Proceedings of The 20th {SIGNLL} Conference on Computational
  Natural Language Learning}, pages 280--290, Berlin, Germany. Association for
  Computational Linguistics.

\bibitem[{Narayan et~al.(2018)Narayan, Cohen, and
  Lapata}]{narayan-etal-2018-dont}
Shashi Narayan, Shay~B. Cohen, and Mirella Lapata. 2018.
\newblock \href {https://doi.org/10.18653/v1/D18-1206} {Don{'}t give me the
  details, just the summary! topic-aware convolutional neural networks for
  extreme summarization}.
\newblock In \emph{Proceedings of the 2018 Conference on Empirical Methods in
  Natural Language Processing}, pages 1797--1807, Brussels, Belgium.
  Association for Computational Linguistics.

\bibitem[{Norouzi et~al.(2016)Norouzi, Bengio, Chen, Jaitly, Schuster, Wu, and
  Schuurmans}]{NIPS2016_2f885d0f}
Mohammad Norouzi, Samy Bengio, zhifeng Chen, Navdeep Jaitly, Mike Schuster,
  Yonghui Wu, and Dale Schuurmans. 2016.
\newblock \href
  {https://proceedings.neurips.cc/paper/2016/file/2f885d0fbe2e131bfc9d98363e55d1d4-Paper.pdf}
  {Reward augmented maximum likelihood for neural structured prediction}.
\newblock In \emph{Advances in Neural Information Processing Systems},
  volume~29, pages 1723--1731. Curran Associates, Inc.

\bibitem[{Och(2003)}]{och-2003-minimum}
Franz~Josef Och. 2003.
\newblock \href {https://doi.org/10.3115/1075096.1075117} {Minimum error rate
  training in statistical machine translation}.
\newblock In \emph{Proceedings of the 41st Annual Meeting of the Association
  for Computational Linguistics}, pages 160--167, Sapporo, Japan. Association
  for Computational Linguistics.

\bibitem[{Och et~al.(2004)Och, Gildea, Khudanpur, Sarkar, Yamada, Fraser,
  Kumar, Shen, Smith, Eng, Jain, Jin, and Radev}]{och-etal-2004-smorgasbord}
Franz~Josef Och, Daniel Gildea, Sanjeev Khudanpur, Anoop Sarkar, Kenji Yamada,
  Alex Fraser, Shankar Kumar, Libin Shen, David Smith, Katherine Eng, Viren
  Jain, Zhen Jin, and Dragomir Radev. 2004.
\newblock \href {https://aclanthology.org/N04-1021} {A smorgasbord of features
  for statistical machine translation}.
\newblock In \emph{Proceedings of the Human Language Technology Conference of
  the North {A}merican Chapter of the Association for Computational
  Linguistics: {HLT}-{NAACL} 2004}, pages 161--168, Boston, Massachusetts, USA.
  Association for Computational Linguistics.

\bibitem[{Pan et~al.(2021)Pan, Wang, Wu, and Li}]{pan-etal-2021-contrastive}
Xiao Pan, Mingxuan Wang, Liwei Wu, and Lei Li. 2021.
\newblock \href {https://doi.org/10.18653/v1/2021.acl-long.21} {Contrastive
  learning for many-to-many multilingual neural machine translation}.
\newblock In \emph{Proceedings of the 59th Annual Meeting of the Association
  for Computational Linguistics and the 11th International Joint Conference on
  Natural Language Processing (Volume 1: Long Papers)}, pages 244--258, Online.
  Association for Computational Linguistics.

\bibitem[{Pang and He(2021)}]{pang2021text}
Richard~Yuanzhe Pang and He~He. 2021.
\newblock \href {https://openreview.net/forum?id=RovX-uQ1Hua} {Text generation
  by learning from demonstrations}.
\newblock In \emph{International Conference on Learning Representations}.

\bibitem[{Paulus et~al.(2018)Paulus, Xiong, and Socher}]{paulus2018a}
Romain Paulus, Caiming Xiong, and Richard Socher. 2018.
\newblock \href {https://openreview.net/forum?id=HkAClQgA-} {A deep reinforced
  model for abstractive summarization}.
\newblock In \emph{International Conference on Learning Representations}.

\bibitem[{Raffel et~al.(2020)Raffel, Shazeer, Roberts, Lee, Narang, Matena,
  Zhou, Li, and Liu}]{JMLR:v21:20-074}
Colin Raffel, Noam Shazeer, Adam Roberts, Katherine Lee, Sharan Narang, Michael
  Matena, Yanqi Zhou, Wei Li, and Peter~J. Liu. 2020.
\newblock \href {http://jmlr.org/papers/v21/20-074.html} {Exploring the limits
  of transfer learning with a unified text-to-text transformer}.
\newblock \emph{Journal of Machine Learning Research}, 21(140):1--67.

\bibitem[{Ranzato et~al.(2016)Ranzato, Chopra, Auli, and
  Zaremba}]{DBLP:journals/corr/RanzatoCAZ15}
Marc'Aurelio Ranzato, Sumit Chopra, Michael Auli, and Wojciech Zaremba. 2016.
\newblock \href {http://arxiv.org/abs/1511.06732} {Sequence level training with
  recurrent neural networks}.
\newblock In \emph{4th International Conference on Learning Representations,
  {ICLR} 2016, San Juan, Puerto Rico, May 2-4, 2016, Conference Track
  Proceedings}.

\bibitem[{Rush et~al.(2015)Rush, Chopra, and Weston}]{rush-etal-2015-neural}
Alexander~M. Rush, Sumit Chopra, and Jason Weston. 2015.
\newblock \href {https://doi.org/10.18653/v1/D15-1044} {A neural attention
  model for abstractive sentence summarization}.
\newblock In \emph{Proceedings of the 2015 Conference on Empirical Methods in
  Natural Language Processing}, pages 379--389, Lisbon, Portugal. Association
  for Computational Linguistics.

\bibitem[{Sandhaus(2008)}]{linguistic2008new}
Evan Sandhaus. 2008.
\newblock \href {https://catalog.ldc.upenn.edu/LDC2008T19} {\emph{The New York
  Times Annotated Corpus}}.
\newblock LDC corpora. Linguistic Data Consortium.

\bibitem[{Schick and Sch{\"u}tze(2021)}]{schick-schutze-2021-shot}
Timo Schick and Hinrich Sch{\"u}tze. 2021.
\newblock \href {https://doi.org/10.18653/v1/2021.emnlp-main.32} {Few-shot text
  generation with natural language instructions}.
\newblock In \emph{Proceedings of the 2021 Conference on Empirical Methods in
  Natural Language Processing}, pages 390--402, Online and Punta Cana,
  Dominican Republic. Association for Computational Linguistics.

\bibitem[{Shen et~al.(2004)Shen, Sarkar, and
  Och}]{shen-etal-2004-discriminative}
Libin Shen, Anoop Sarkar, and Franz~Josef Och. 2004.
\newblock \href {https://aclanthology.org/N04-1023} {Discriminative reranking
  for machine translation}.
\newblock In \emph{Proceedings of the Human Language Technology Conference of
  the North {A}merican Chapter of the Association for Computational
  Linguistics: {HLT}-{NAACL} 2004}, pages 177--184, Boston, Massachusetts, USA.
  Association for Computational Linguistics.

\bibitem[{Shen et~al.(2016)Shen, Cheng, He, He, Wu, Sun, and
  Liu}]{shen-etal-2016-minimum}
Shiqi Shen, Yong Cheng, Zhongjun He, Wei He, Hua Wu, Maosong Sun, and Yang Liu.
  2016.
\newblock \href {https://doi.org/10.18653/v1/P16-1159} {Minimum risk training
  for neural machine translation}.
\newblock In \emph{Proceedings of the 54th Annual Meeting of the Association
  for Computational Linguistics (Volume 1: Long Papers)}, pages 1683--1692,
  Berlin, Germany. Association for Computational Linguistics.

\bibitem[{Stahlberg and Byrne(2019)}]{stahlberg-byrne-2019-nmt}
Felix Stahlberg and Bill Byrne. 2019.
\newblock \href {https://doi.org/10.18653/v1/D19-1331} {On {NMT} search errors
  and model errors: Cat got your tongue?}
\newblock In \emph{Proceedings of the 2019 Conference on Empirical Methods in
  Natural Language Processing and the 9th International Joint Conference on
  Natural Language Processing (EMNLP-IJCNLP)}, pages 3356--3362, Hong Kong,
  China. Association for Computational Linguistics.

\bibitem[{Sun and Li(2021)}]{DBLP:journals/corr/abs-2108-11846}
Shichao Sun and Wenjie Li. 2021.
\newblock \href {http://arxiv.org/abs/2108.11846} {Alleviating exposure bias
  via contrastive learning for abstractive text summarization}.
\newblock \emph{CoRR}, abs/2108.11846.

\bibitem[{Sutskever et~al.(2014)Sutskever, Vinyals, and
  Le}]{10.5555/2969033.2969173}
Ilya Sutskever, Oriol Vinyals, and Quoc~V. Le. 2014.
\newblock Sequence to sequence learning with neural networks.
\newblock In \emph{Proceedings of the 27th International Conference on Neural
  Information Processing Systems - Volume 2}, NIPS'14, page 3104–3112,
  Cambridge, MA, USA. MIT Press.

\bibitem[{Szegedy et~al.(2016)Szegedy, Vanhoucke, Ioffe, Shlens, and
  Wojna}]{7780677}
C.~Szegedy, V.~Vanhoucke, S.~Ioffe, J.~Shlens, and Z.~Wojna. 2016.
\newblock \href {https://doi.org/10.1109/CVPR.2016.308} {Rethinking the
  inception architecture for computer vision}.
\newblock In \emph{2016 IEEE Conference on Computer Vision and Pattern
  Recognition (CVPR)}, pages 2818--2826, Los Alamitos, CA, USA. IEEE Computer
  Society.

\bibitem[{Taskar et~al.(2004)Taskar, Guestrin, and Koller}]{NIPS2003_878d5691}
Ben Taskar, Carlos Guestrin, and Daphne Koller. 2004.
\newblock \href
  {https://proceedings.neurips.cc/paper/2003/file/878d5691c824ee2aaf770f7d36c151d6-Paper.pdf}
  {Max-margin markov networks}.
\newblock In \emph{Advances in Neural Information Processing Systems},
  volume~16. MIT Press.

\bibitem[{Uehara et~al.(2020)Uehara, Ishigaki, Aoki, Noji, Goshima, Kobayashi,
  Takamura, and Miyao}]{uehara-etal-2020-learning}
Yui Uehara, Tatsuya Ishigaki, Kasumi Aoki, Hiroshi Noji, Keiichi Goshima,
  Ichiro Kobayashi, Hiroya Takamura, and Yusuke Miyao. 2020.
\newblock \href {https://doi.org/10.18653/v1/2020.coling-main.213} {Learning
  with contrastive examples for data-to-text generation}.
\newblock In \emph{Proceedings of the 28th International Conference on
  Computational Linguistics}, pages 2352--2362, Barcelona, Spain (Online).
  International Committee on Computational Linguistics.

\bibitem[{Vijayakumar et~al.(2018)Vijayakumar, Cogswell, Selvaraju, Sun, Lee,
  Crandall, and
  Batra}]{Vijayakumar_Cogswell_Selvaraju_Sun_Lee_Crandall_Batra_2018}
Ashwin Vijayakumar, Michael Cogswell, Ramprasaath Selvaraju, Qing Sun, Stefan
  Lee, David Crandall, and Dhruv Batra. 2018.
\newblock \href {https://ojs.aaai.org/index.php/AAAI/article/view/12340}
  {Diverse beam search for improved description of complex scenes}.
\newblock \emph{Proceedings of the AAAI Conference on Artificial Intelligence},
  32(1).

\bibitem[{Wan et~al.(2015)Wan, Cao, Wei, Li, and Zhou}]{Wan2015MultiDocumentSV}
Xiaojun Wan, Ziqiang Cao, Furu Wei, Sujian Li, and M.~Zhou. 2015.
\newblock Multi-document summarization via discriminative summary reranking.
\newblock \emph{ArXiv}, abs/1507.02062.

\bibitem[{Wang et~al.(2020)Wang, Tu, Shi, and Liu}]{wang-etal-2020-inference}
Shuo Wang, Zhaopeng Tu, Shuming Shi, and Yang Liu. 2020.
\newblock \href {https://doi.org/10.18653/v1/2020.acl-main.278} {On the
  inference calibration of neural machine translation}.
\newblock In \emph{Proceedings of the 58th Annual Meeting of the Association
  for Computational Linguistics}, pages 3070--3079, Online. Association for
  Computational Linguistics.

\bibitem[{Wieting et~al.(2019)Wieting, Berg-Kirkpatrick, Gimpel, and
  Neubig}]{wieting-etal-2019-beyond}
John Wieting, Taylor Berg-Kirkpatrick, Kevin Gimpel, and Graham Neubig. 2019.
\newblock \href {https://doi.org/10.18653/v1/P19-1427} {Beyond {BLEU}:training
  neural machine translation with semantic similarity}.
\newblock In \emph{Proceedings of the 57th Annual Meeting of the Association
  for Computational Linguistics}, pages 4344--4355, Florence, Italy.
  Association for Computational Linguistics.

\bibitem[{Wiseman and Rush(2016)}]{wiseman-rush-2016-sequence}
Sam Wiseman and Alexander~M. Rush. 2016.
\newblock \href {https://doi.org/10.18653/v1/D16-1137} {Sequence-to-sequence
  learning as beam-search optimization}.
\newblock In \emph{Proceedings of the 2016 Conference on Empirical Methods in
  Natural Language Processing}, pages 1296--1306, Austin, Texas. Association
  for Computational Linguistics.

\bibitem[{Wolf et~al.(2020)Wolf, Debut, Sanh, Chaumond, Delangue, Moi, Cistac,
  Rault, Louf, Funtowicz, Davison, Shleifer, von Platen, Ma, Jernite, Plu, Xu,
  Le~Scao, Gugger, Drame, Lhoest, and Rush}]{wolf-etal-2020-transformers}
Thomas Wolf, Lysandre Debut, Victor Sanh, Julien Chaumond, Clement Delangue,
  Anthony Moi, Pierric Cistac, Tim Rault, Remi Louf, Morgan Funtowicz, Joe
  Davison, Sam Shleifer, Patrick von Platen, Clara Ma, Yacine Jernite, Julien
  Plu, Canwen Xu, Teven Le~Scao, Sylvain Gugger, Mariama Drame, Quentin Lhoest,
  and Alexander Rush. 2020.
\newblock \href {https://doi.org/10.18653/v1/2020.emnlp-demos.6} {Transformers:
  State-of-the-art natural language processing}.
\newblock In \emph{Proceedings of the 2020 Conference on Empirical Methods in
  Natural Language Processing: System Demonstrations}, pages 38--45, Online.
  Association for Computational Linguistics.

\bibitem[{Xu et~al.(2021)Xu, Zhang, Wu, and
  Wei}]{DBLP:journals/corr/abs-2109-03481}
Shusheng Xu, Xingxing Zhang, Yi~Wu, and Furu Wei. 2021.
\newblock \href {http://arxiv.org/abs/2109.03481} {Sequence level contrastive
  learning for text summarization}.
\newblock \emph{CoRR}, abs/2109.03481.

\bibitem[{Yang et~al.(2019)Yang, Cheng, Liu, and Sun}]{yang-etal-2019-reducing}
Zonghan Yang, Yong Cheng, Yang Liu, and Maosong Sun. 2019.
\newblock \href {https://doi.org/10.18653/v1/P19-1623} {Reducing word omission
  errors in neural machine translation: A contrastive learning approach}.
\newblock In \emph{Proceedings of the 57th Annual Meeting of the Association
  for Computational Linguistics}, pages 6191--6196, Florence, Italy.
  Association for Computational Linguistics.

\bibitem[{Yuan et~al.(2021)Yuan, Neubig, and Liu}]{yuan2021bartscore}
Weizhe Yuan, Graham Neubig, and Pengfei Liu. 2021.
\newblock \href {https://openreview.net/forum?id=5Ya8PbvpZ9} {{BARTS}core:
  Evaluating generated text as text generation}.
\newblock In \emph{Thirty-Fifth Conference on Neural Information Processing
  Systems}.

\bibitem[{Zhang et~al.(2020)Zhang, Zhao, Saleh, and Liu}]{zhang2020pegasus}
Jingqing Zhang, Yao Zhao, Mohammad Saleh, and Peter Liu. 2020.
\newblock Pegasus: Pre-training with extracted gap-sentences for abstractive
  summarization.
\newblock In \emph{International Conference on Machine Learning}, pages
  11328--11339. PMLR.

\bibitem[{Zhang* et~al.(2020)Zhang*, Kishore*, Wu*, Weinberger, and
  Artzi}]{Zhang*2020BERTScore:}
Tianyi Zhang*, Varsha Kishore*, Felix Wu*, Kilian~Q. Weinberger, and Yoav
  Artzi. 2020.
\newblock \href {https://openreview.net/forum?id=SkeHuCVFDr} {Bertscore:
  Evaluating text generation with bert}.
\newblock In \emph{International Conference on Learning Representations}.

\bibitem[{Zhang et~al.(2019)Zhang, Feng, Meng, You, and
  Liu}]{zhang-etal-2019-bridging}
Wen Zhang, Yang Feng, Fandong Meng, Di~You, and Qun Liu. 2019.
\newblock \href {https://doi.org/10.18653/v1/P19-1426} {Bridging the gap
  between training and inference for neural machine translation}.
\newblock In \emph{Proceedings of the 57th Annual Meeting of the Association
  for Computational Linguistics}, pages 4334--4343, Florence, Italy.
  Association for Computational Linguistics.

\bibitem[{Zhong et~al.(2020)Zhong, Liu, Chen, Wang, Qiu, and
  Huang}]{zhong-etal-2020-extractive}
Ming Zhong, Pengfei Liu, Yiran Chen, Danqing Wang, Xipeng Qiu, and Xuanjing
  Huang. 2020.
\newblock \href {https://doi.org/10.18653/v1/2020.acl-main.552} {Extractive
  summarization as text matching}.
\newblock In \emph{Proceedings of the 58th Annual Meeting of the Association
  for Computational Linguistics}, pages 6197--6208, Online. Association for
  Computational Linguistics.

\end{thebibliography}
\bibliographystyle{acl_natbib}

\appendix

\newpage

\section{Datasets Statistics}

\label{sec:datasets}
\begin{table}[h]
  \centering
  \small
    \begin{tabular}{@{\extracolsep{1pt}}lccccc}
    \toprule
    \multirow{2}{*}{Datasets} & \multicolumn{3}{c}{\# Examples} & \multicolumn{2}{c}{Avg. Words} \\
    \cmidrule{2-4} \cmidrule{5-6}
    & Train & Valid & Test & Doc. & Sum. \\
    \midrule
    CNNDM & 287K & 13K & 11K & 791.6 & 55.6 \\
    XSum & 203K & 11K & 11K & 429.2 & 23.3 \\
    NYT & 44K & 5.5K & 6.4K & 1320.2 & 123.4\\
    \bottomrule
    \end{tabular}%
  \caption{Datasets Statistics.}
  \label{tab:data}%
\end{table}%

\section{Implementation Details}
\label{sec:appendix}
We use diverse beam search~\citep{Vijayakumar_Cogswell_Selvaraju_Sun_Lee_Crandall_Batra_2018} to generate 16 candidates for each data sample.
On \texttt{CNNDM} and \texttt{XSum}, we use the pre-trained BART\footnote{The checkpoint is ``facebook/bart-large-cnn'', containing around 400M parameters.} and PEGASUS\footnote{The checkpoint is ``google/pegasus-xsum"" containing around 568M parameters.} models from the \textit{Transformers}~\citep{wolf-etal-2020-transformers} library as the base abstractive models for candidate summary generation and model finetuning respectively.
On \texttt{NYT}, we first fine-tuned a BART model\footnote{The checkpoint is ``facebook/bart-large''.} with MLE training as the base abstractive model, since our data pre-processing is sightly different from the previous work and there are no available pre-trained checkpoints.  
We use 4 NVIDIA RTX 3090 GPUs for the model training, and the average running time for one epoch is around 20 hours.
We use the Adam optimizer~\citep{DBLP:journals/corr/KingmaB14} with learning rate scheduling for the model training:
\begin{equation}
\small
    lr = \num{2e-3}\min(\textrm{step}^{-0.5}, \textrm{step}\cdot\textrm{warmup}^{-1.5}) \nonumber
\end{equation}
where $\mathrm{warmup}$ denotes the warmup steps, which is set to 10000, $\mathrm{step}$ is the number of updating steps, $lr$ is the learning rate.

We set the length penalty factor $\alpha$ in the scoring function (Eq.~\ref{eq:score}) to the same value as used in the original beam search.
We search the value of the margin $\lambda$ in the contrastive loss (Eq.~\ref{eq:contr}) within the range 
$[\num{1e-5}, 1]$,
and decide the value based on the model performance on the validation set.
We also performed extensive search for the coefficient $\gamma$ in Eq.~\ref{eq:mul}.
The specific hyper-parameter setting is reported in Tab.~\ref{tab:appendix}.

We use the standard ROUGE~\citep{lin-2004-rouge} Perl package\footnote{\url{https://github.com/summanlp/evaluation/tree/master/ROUGE-RELEASE-1.5.5}} for evaluation.
The command line parameters are `-c 95 -r 1000 -n 2 -m'.
Before the ROUGE evaluation, the reference summaries and system outputs are lower-cased and tokenized.\footnote{PTB tokenizer is used for tokenization. \url{https://nlp.stanford.edu/nlp/javadoc/javanlp/edu/stanford/nlp/process/PTBTokenizer.html}}

\begin{table}[t]
\centering
\begin{tabular}{lccc}
\toprule
\textbf{Datasets} &  $\mathbf{\lambda}$ (Eq.~\ref{eq:contr}) & $\mathbf{\alpha}$ (Eq.~\ref{eq:score})   &  $\mathbf{\gamma}$ (Eq.~\ref{eq:mul}) \\
\midrule
 CNNDM & 0.001 & 2.0 & 100 \\
 XSum & 0.1 & 0.6 & 100 \\
 NYT & 0.001 & 2.0 & 100  \\
\bottomrule
\end{tabular}
\caption{\label{tab:appendix} Hyper-parameter Setting.}
\end{table}

\section{Details of Few-shot Fine-tuning}
\label{appendix:few-shot}

On \texttt{CNNDM}, we randomly select 100 examples from the training set for fine-tuning.
On \texttt{XSum}, we found that at least 1000 examples are needed for the model to achieve better performance compared to the baseline model.
All experiments are repeated three times.
We randomly select 1000 examples from the original validation set for hyper-parameter selection.
We use the Adam optimizer with the learning rate set to $\num{1e-6}$.
The model is trained for 15 epochs on \texttt{CNNDM} and 10 epochs on \texttt{XSum}.


\end{document}